\documentclass[10pt, a4paper]{article}
\usepackage[]{lrec-coling2024} 

\usepackage{pdfpages}
\usepackage{multirow}
\usepackage{multicol}
\usepackage{enumitem,kantlipsum}
\usepackage[normalem]{ulem}
\usepackage{color,colortbl}
\usepackage{todonotes} 
\usepackage{soul}
 \usepackage{threeparttable, makecell}

\usepackage{booktabs} 
\usepackage{textalpha} 

\newcommand{\rashkincorpus}{\texttt{\textbf{TSHP-17}}}
\newcommand{\proppycorpus}{\texttt{\textbf{QProp}} }

\usepackage{makecell}
\usepackage{xcolor}

\usepackage{graphicx}

\usepackage{color}

\newcommand{\arpro}{\textbf{\textit{ArPro}}}

\title{Can GPT-4 Identify Propaganda? Annotation and Detection of Propaganda Spans in News Articles}

\name{Maram Hasanain,  Fatema Ahmed, Firoj Alam} 

\address{Qatar Computing Research Institute, HBKU, Qatar \\          \{mhasanain,fakter,fialam\}@hbku.edu.qa\\
}

\abstract{
The use of propaganda has spiked on mainstream and social media, aiming to manipulate or mislead users. While efforts to automatically detect propaganda techniques in textual, visual, or multimodal content have increased, most of them primarily focus on English content. The majority of the recent initiatives targeting medium to low-resource languages produced relatively small annotated datasets, with a skewed distribution, posing challenges for the development of sophisticated propaganda detection models. To address this challenge, we carefully develop the largest propaganda dataset to date, \arpro{}, comprised of $8K$ paragraphs from newspaper articles, labeled at the text span level following a taxonomy of 23 propagandistic techniques. Furthermore, our work offers the \textit{first attempt} to understand the performance of large language models (LLMs), using GPT-4, for fine-grained propaganda detection from text. Results showed that GPT-4's performance degrades as the task moves from simply classifying a paragraph as propagandistic or not, to the fine-grained task of detecting propaganda techniques and their manifestation in text. Compared to models fine-tuned on the dataset for propaganda detection at different classification granularities, GPT-4 is still far behind. Finally, we evaluate GPT-4 on a dataset consisting of six other languages for span detection, and results suggest that the model struggles with the task across languages. Our dataset and resources will be released to the community.
\\ \newline  
}

\begin{document}

\maketitleabstract

\section{Introduction}
\label{sec:introduction}

Online media has become a primary channel for information dissemination and consumption, with numerous individuals considering it their main source of news~\cite{perrin2015social}. While online media (including news and social media platforms) offers a plethora of benefits, it is periodically exploited by malicious actors aiming to manipulate and mislead a broad audience. They often engage in sharing inappropriate content, misinformation, and disinformation \citep{alam-etal-2022-survey,ijcai2022p781}. Among the various forms of misleading and harmful content, propaganda is another communication tool designed to influence opinions and actions to achieve a specific objective~\cite{InstituteforPropagandaAnalysis1938}.

To combat the proliferation of propaganda on online platforms computationally, there has been a significant surge in research in recent years. The aim is to automatically identify propagandistic content in textual, visual, and multimodal modalities, such as memes \cite{chen2023multimodal,SemEval2021-6-Dimitrov,EMNLP19DaSanMartino}. Initial studies on propaganda detection primarily focused on binary classification (differentiating between propagandistic and non-propagandistic text) and multiclass classification~\cite{BARRONCEDENO20191849,rashkin-EtAl:2017:EMNLP2017}. Building on this foundation, \citet{EMNLP19DaSanMartino} curated a list of propagandistic techniques used in text to sway readers opinions and actions, such as \textit{name calling}, \textit{appeal to fear}, \textit{misrepresentation of someone's opinion (straw man fallacy)}, and \textit{causal oversimplification}. Such efforts have paved the way for the creation of significant resources, primarily in the English language. A recent effort has expanded the prior work to include multilingual propaganda detection~\cite{piskorski2023semeval}.

\begin{figure}
    \centering
    \includegraphics[width=1\linewidth]{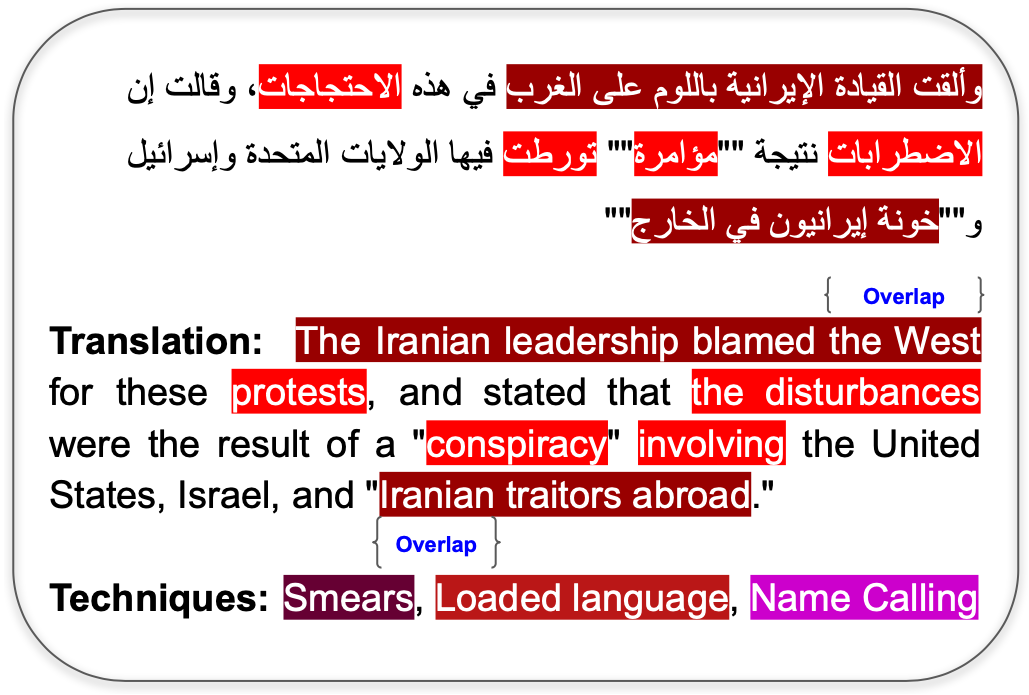}
    \vspace{-0.2cm}
    \caption{An example of a news paragraph annotated with propagandistic techniques. The words ``traitor'' and ``the West'' were labeled as \textit{name calling}, which is not visible due to \textcolor{blue}{overlap} with other techniques.}
    \label{fig:prop_example}
    \vspace{-0.5cm}
\end{figure}
%

Research on Arabic content is relatively sparse. Previous datasets, such as the one proposed in prior studies, have been primarily focused on tweets and are notably limited in size~\cite{propaganda-detection:WANLP2022-overview}. Thus, our work aims to develop a more comprehensive annotated dataset including annotations at the span level. We annotated a large set of news paragraphs collected from Arabic news articles. Annotating text with propagandistic techniques is inherently complex due to subjectivity, contextual variations, linguistic and cultural nuances, and cognitive biases. In Figure \ref{fig:prop_example}, we present a paragraph annotated at the text span level with various propaganda techniques. Techniques can be even overlapping in the same text spans. More details are discussed in Section \ref{ssec:data_annotation}.

Several recent studies have shown that current LLMs (e.g., ChatGPT) can be effectively employed for downstream NLP tasks, and they have found that the performance gap between state-of-the-art (SOTA) methods and GPTs (e.g., GPT-4) is relatively small \cite{bang2023multitask,ahuja2023mega,abdelali2023benchmarking,liang2022holistic}. Inspired by these studies, we aim to leverage GPT-4~\cite{openai2023gpt4} for propaganda detection across various granularities, ranging from binary classification to span detection with propagandistic techniques. We compare the performance of the model with different fine-tuned models. 



To summarize, our contributions are three-fold: 
\begin{itemize}[itemsep=0pt, parsep=0pt]
    \item We investigate the performance of GPT-4 for propagandistic techniques span detection and labeling. This is the \textit{first attempt} for this task. GPT-4's performance is also compared to several fine-tuned models.
    \item We release the \textit{largest dataset to date}, named \textit{\textbf{\arpro}}, for fine-grained propaganda detection\footnote{\url{https://www.anonymous.com}}, in addition to the associated extensive Arabic annotation guidelines for the task. 
    \item We provide detailed insights into the data collection and annotation process, as well as comprehensive statistics on the dataset.
\end{itemize}

The main findings are as follows: 
\emph{(i)} Span-level propagandistic techniques annotation is a complex process, and a two-step annotation approach leads to improved annotation agreement; 
\emph{(ii)} The distribution of some techniques is skewed, corroborating previous findings~\cite{propaganda-detection:WANLP2022-overview}, which requires further studies to understand whether such techniques are inherently scarcely used in news reporting; 
\emph{(iii)} Fine-tuned models consistently outperform GPT-4 in a zero-shot setting; 
\emph{(iv)} GPT-4 consistently fails to detect span-level propagandistic techniques in a zero-shot setting across multiple languages.


\section{\arpro{} Dataset}
\label{sec:dataset}
Our dataset is constructed to be the largest dataset for the task, in scale of or larger than datasets in multiple languages~\cite{piskorski2023news}. The process included three steps as detailed in the following sections: (1) acquiring raw data, (2) preparing and sampling data for annotation, and (3) the manual annotation phase.
\subsection{Data Collection}
We decided to annotate news articles from a variety of Arabic news domains. Our dataset is based on two collections of such articles: \emph{(i)} AraFacts, and \emph{(ii)} a large-scale in-house collection. 
The \textbf{AraFacts} dataset~\cite{ali2021arafacts} contains true and false Arabic claims verified by fact-checking websites, and each claim is associated with online sources propagating or negating the claim. We only keep Web pages that are from news domains in the set (e.g., www.alquds.co.uk). 
Since fake news can be used for propaganda purposes~\cite{vamanu-2019}, we hypothesized that such news articles discussing controversial claims have a higher chance of containing propaganda.

As for our \textbf{in-house dataset}, it consists of $600K$ news articles from over 400 news domains with articles being as old as January 2013 and up to January 2023. This dataset offers versatility and wide coverage of Arabic news websites, allowing our final annotated dataset to be representative of a variety of writing styles and topics. 

\begin{figure*}
    \centering
    \includegraphics[width=1\linewidth]{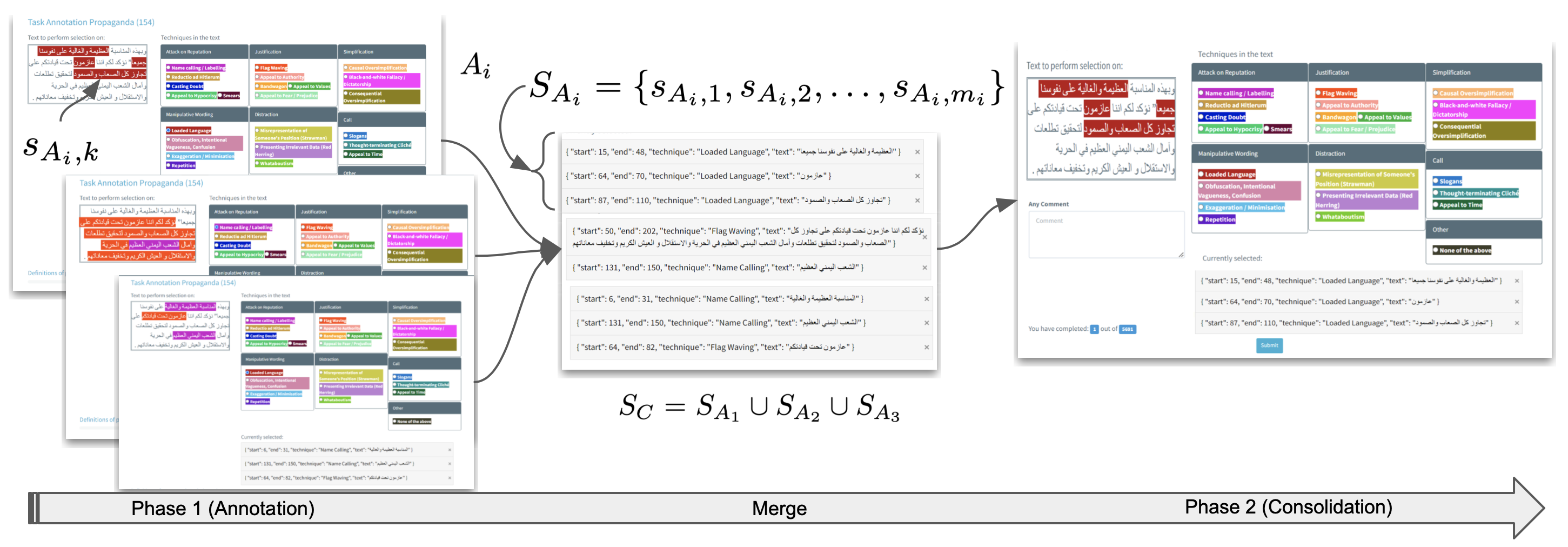}
    \vspace{-0.5cm}
    \caption{The entire annotation process.}
    \label{fig:annotation_process}
    \vspace{-0.4cm}
\end{figure*}

\noindent\textbf{Data Preparation.}
We automatically parsed the news articles using existing Python packages: Goose3,\footnote{\url{https://goose3.readthedocs.io/en/latest/quickstart.html}} Newspaper3k,\footnote{\url{https://newspaper.readthedocs.io/en/latest/}} and Trafilatura.\footnote{\url{https://trafilatura.readthedocs.io/en}} In our experience, we found that none of these popular parsing packages lead to optimal results. Thus, we apply all three parsing packages to all articles, and for each article, we opted to select the longest extracted content. This setup yielded better extraction performance when manually evaluated over a sample of the articles. The parsed articles were then split into paragraphs, assuming paragraphs are those separated by a blank line. This resulted in over $4M$ paragraphs. 

\noindent\textbf{Paragraph Selection.} We applied a thorough filter over the paragraphs to overcome problems due to automatic parsing, by filtering out paragraphs matching any of the following: (i) non-Arabic as classified by langdetect,\footnote{\url{https://pypi.org/project/langdetect/}} (ii) containing HTML tags, and (iii) containing any special character repeated more than three times (e.g., \%, *, etc.). The paragraphs were de-duplicated using Cosine similarity, with a similarity $>=0.75$ indicating duplication. Finally, we only keep paragraphs that have at least 14 words and shorter than 78 words. This length range is based on the length distribution of all paragraphs, as most paragraphs fall within this range. The resulting set included $2.65M$ paragraphs. 

We construct the final set of paragraphs to annotate as follows. Paragraphs of full articles sourced from AraFacts were included. This is to allow article-level analysis and detection of propaganda over \arpro. We apply stratified sampling by randomly sampling 30 or less paragraphs per news domain from articles of the in-house dataset.\footnote{Majority of news domains has 30 or more paragraphs, thus, we use 30 as our cutoff.} This guarantees coverage of a large and versatile set of news websites. 



\subsection{Data Annotation}
\label{ssec:data_annotation}
The paragraphs were annotated adopting an existing two-tier taxonomy of \textbf{six} main categories, grouping \textbf{23} persuasion techniques~\cite{piskorski2023news}. That taxonomy represents the most comprehensive effort in literature aiming to model fine-grained propaganda use in text over a variety of languages. Annotation was guided by \textit{Arabic} annotation guidelines we created for the task, inspired by English  guidelines developed by~\citet{piskorski2023news}. In our prior work,\footnote{Reference removed for anonymity} we observed that annotation guidelines in the same language of the data are crucial, not only to capture linguistic nuances, but also to make the annotation process more convenient for annotators. The guidelines included several examples of paragraphs per technique, sourced from existing Arabic news articles. It was reviewed by several NLP experts who are also native Arabic speakers. The annotation guidelines can be found in Appendix D. 
 Since a text span may have multiple techniques, annotators were instructed to annotate text that can overlap, as shown in Figure~\ref{fig:prop_example}. The techniques \textit{smears} and \textit{name calling and labeling} overlapped in a text span. 
Our annotation process includes two phases: 
\begin{itemize}[noitemsep,nolistsep]
    \item \textbf{Phase 1 (\textit{annotation}):} In this phase, each paragraph was presented to three annotators. They were instructed to identify the 23 propaganda techniques in the text and to highlight the corresponding text span for each label. 
    \item \textbf{Phase 2 (\textit{consolidation}):} Annotations from Phase 1, were presented to two expert annotators (referred to as consolidators). 
    The purpose of this phase was to review the annotations and resolve any disagreements. To maintain the quality of the annotations, we arranged for two consolidators to collaboratively review the work. Moreover, they were also encouraged to 
    identify techniques that the initial annotators might have missed. Consolidators were requested to also give their general observation on the quality of the data which served as additional training for both teams and a resource for improving the training process for future tasks. 
    This phase resulted in the final gold annotations.     
\end{itemize}

\noindent\textbf{Annotation platform.} We utilized our in-house annotation platform for the annotation task. Separate annotation interfaces were designed for each phase.


\noindent\textbf{Annotation Process}
Figure \ref{fig:annotation_process} summarizes the entire annotation process, from phase 1 to phase 2. The process is formulated as follows: 
Let us consider each annotator $A_i$ provides a set of spans $S_{A_i}$ and each span in $ S_{A_i} $ is represented as $ s_{A_i,k} $, where $ k $ is the index of the span for the $ i $-th annotator. Note that $ k $ can range from 1 to the total number of spans identified by annotator $A_i$, and this total can be different for each annotator. Given this representation, for the $i^{th}$ annotator the set of spans is defined as
$S_{A_i} = \{ s_{A_i,1}, s_{A_i,2}, \dots, s_{A_i,m_i} \}$
where $m_i$ is the total number of spans identified by annotator $A_i$. To combine the spans of all annotators, we formed the union of their span sets as $S_C = S_{A_1} \cup S_{A_2} \cup S_{A_3}$. The combined set $S_C$  will contain all unique spans identified by all annotators. This set goes through the consolidation phase to finalize the annotations by consolidators.


The annotation guidelines allow grouping techniques as shown in Figure~\ref{fig:annotation_process}, consisting of six main categories. We included ``Other'' to account for cases when a propagandistic text span fits none of the categories. 

\noindent\textbf{Annotation Team.} The team in phase 1 consisted of seven members. The consolidation team included four members;  two of them had prior experience working as annotators for various tasks in Arabic. The consolidation team's members also had prior experience with Arabic NLP. All 
annotators are native Arabic speakers, holding at least a bachelor's degree, with two members holding a Ph.D. Both teams were provided with written guidelines and received several rounds of training. More details in Appendix~\ref{sec:app_annotation_training}. 
 Both teams were supervised, monitored, and trained by an expert annotator, who also handled quality control throughout the entire annotation process. This quality assurance included periodic checks of random annotation samples and giving feedback to both teams.

\subsection{Inter-Annotation Agreement}
\label{ssec:anno_agr}
The span-level annotation process is a very complex task. The subjective nature of the task adds more complexity to the annotation process. We computed annotation agreement considering different settings following prior studies: \emph{(i)} multiclass multilabel for the paragraphs \cite{dimitrov2021detecting}, \emph{(ii)} 
binary labels (containing or not containing propaganda techniques in the paragraphs), and \emph{(iii)} span labels \cite{EMNLP19DaSanMartino}. 

For the multiclass multilabel and binary settings, we computed Krippendorff's $\alpha$, which is suitable for such agreement computation \cite{artstein2008inter,passonneau2006measuring}. This was calculated between each annotator and the consolidated label for the entire dataset, comprising 8,000 paragraphs. This yielded an average Krippendorff's $\alpha$ value of 0.335, as shown in Table \ref{tab:annotation_agreement}.\footnote{Our agreement is significantly higher than that for related tasks \cite{Roitero2020}: Krippendorff's $\alpha$ ranges between 0.066 and 0.131.} Additionally, we assessed the annotation agreement from phase 1, where each paragraph was annotated by three annotators. This resulted in a Krippendorff's $\alpha$ value of 0.243 and 0.296 for multilabel and binary settings, respectively. The agreement score from phase 1 shows the significance of the second phase of annotation involving expert annotators.

For the span label annotation, we computed $\gamma$ 
\cite{Mathet2015, mathet-2017-agreement}, which have been studied in similar tasks \cite{EMNLP19DaSanMartino}. The $\gamma$ agreement is specifically designed for span/segment level annotation tasks to take into account span boundaries (i.e., start and end) and their labels. Note that it also allows for consideration of overlapping annotations, as depicted in Figure \ref{fig:prop_example}. 
Overall, the span and binary label annotations shows moderate to substantial agreement among annotators between two phases.\footnote{Recall that values of Kappa of 0.21--0.40, 0.41--0.60, 0.61--0.80, and 0.81--1.0 correspond to fair, moderate, substantial and perfect agreement, respectively~\cite{landis1977measurement}.}
 
\begin{table}[]
\centering
\setlength{\tabcolsep}{3pt}
\resizebox{0.8\columnwidth}{!}{%
\begin{tabular}{@{}lrrr@{}}
\toprule
\textbf{Agr. Pair} & \textbf{Multilabel ($\alpha$}) & \textbf{Binary ($\alpha$}) & \textbf{Span ($\gamma$}) \\ \midrule
A 1 - C & 0.598 & 0.810 & 0.714 \\
A 2 - C & 0.448 & 0.697 & 0.258 \\
A 3 - C & 0.420 & 0.668 & 0.604 \\
A 4 - C & 0.351 & 0.637 & 0.397 \\
A 5 - C & 0.308 & 0.541 & 0.651 \\
A 6 - C & 0.270 & 0.537 & 0.653 \\
A 7 - C & 0.234 & 0.461 & 0.510 \\ \midrule
\textbf{Average} & \textbf{0.375} & \textbf{0.622} & \textbf{0.546} \\ \bottomrule
\end{tabular}%
}
\vspace{-0.2cm}
\caption{Annotation agreement among different annotators vs. consolidator in multiclass multilabel, binary and span levels. A: Annotator; C: Consolidator. Agr.: Agreement.}
\label{tab:annotation_agreement}
\vspace{-0.2cm}
\end{table}

\begin{table}[]
\centering
\setlength{\tabcolsep}{3pt}
\resizebox{0.5\columnwidth}{!}{%
\begin{tabular}{@{}lr@{}}
\toprule
\textbf{Content} & \textbf{Stat} \\ \midrule
\# News articles & 2,810 \\
\# Paragraphs & 8,000 \\
\# Sentences & 10,331 \\
\# Words & 277,952 \\
Avg sentence length & 26.90 \\
Avg paragraph length & 34.74 \\ \bottomrule
\end{tabular}%
}
\vspace{-0.2cm}
\caption{Descriptive statistics of the dataset.}
\label{tab:desc_stat}
\vspace{-0.4cm}
\end{table}


\section{Statistics and Analysis}
\label{sec:data_analysis}

\noindent\textbf{Basic Statistics.} Table \ref{tab:desc_stat} summarizes basic statistics of the dataset. In total, the dataset consists of $8K$ annotated paragraphs selected from $2.8K$ news articles; $\sim$10K sentences; and $\sim$277K words. 
\arpro{} covers news 
from 300 news agencies. 

\noindent\textbf{Distribution of Topics.} In Table \ref{tab:topic_coverage}, we report topic-wise coverage and associated number of paragraphs.\footnote{Datasets from which the news articles were sourced included topic assignments.} The dataset covers 14 different topics among them, \textit{news} and \textit{politics} cover more than 50\% of the paragraphs. Considering the fraction of propagandistic paragraphs from all paragraphs per topic, our analysis suggests that propagandistic content is relatively higher in these two topics. 

\noindent\textbf{Distribution of Techniques.} In Table \ref{tab:data_split_tech_dist}, we report distribution of techniques in the whole dataset. \textit{loaded language} and \textit{name calling and labeling} are the most frequent techniques. We introduced ``no\_technique'' as a label for the formulation of binary and multilabel classification settings. The paragraphs with no annotated technique are labeled as ``no\_technique'' indicating a non-propagandistic paragraph. The \textit{loaded language} constitutes 50\% of the identified propaganda spans, which is inline with the 
previous studies~\cite{SemEval2021-6-Dimitrov}.

\begin{table}[]
\centering
\setlength{\tabcolsep}{3pt}
\resizebox{0.65\columnwidth}{!}{%
\begin{tabular}{@{}lr@{}}
\toprule
\textbf{Topic} & \textbf{\# Paragraph} \\ \midrule
News & 2,993 \\
Politics & 2,330 \\
Health & 594 \\
Social & 473 \\
Sports & 403 \\
Miscellaneous & 286 \\
Arts and Culture & 215 \\
Religion & 210 \\
Science and Technology & 175 \\
Entertainment & 134 \\
Business and Economy & 94 \\
Travel & 65 \\
Human Rights & 24 \\
Lifestyle & 4 \\ \midrule
\textbf{Total}  & \textbf{8,000} \\
\bottomrule
\end{tabular}%
}
\vspace{-0.2cm}
\caption{Number of paragraphs per topic.}
\label{tab:topic_coverage}
\vspace{-0.2cm}
\end{table}


\begin{table}[]
\centering
\setlength{\tabcolsep}{4pt}
\resizebox{0.9\columnwidth}{!}{%
\begin{tabular}{lrrr}
\toprule
\textbf{Technique} & \textbf{F} & \textbf{P-R} & \textbf{T-R} \\ \midrule
Appeal\_to\_Authority             & 256   & 0.032 & 0.012 \\
Appeal\_to\_Fear-Prejudice        & 125   & 0.016 & 0.006 \\
Appeal\_to\_Hypocrisy             & 108   & 0.013 & 0.005 \\
Appeal\_to\_Popularity            & 56    & 0.007 & 0.003 \\
Appeal\_to\_Time                  & 70    & 0.009 & 0.003 \\
Appeal\_to\_Values                & 52    & 0.006 & 0.003 \\
Causal\_Oversimplification        & 389   & 0.049 & 0.019 \\
Consequential\_Oversimplification & 110   & 0.014 & 0.005 \\
Conversation\_Killer              & 72    & 0.009 & 0.004 \\
Doubt                             & 303   & 0.038 & 0.015 \\
Exaggeration-Minimisation         & 1,290  & 0.161 & 0.063 \\
False\_Dilemma-No\_Choice         & 79    & 0.010 & 0.004 \\
Flag\_Waving                      & 237   & 0.030 & 0.012 \\
Guilt\_by\_Association            & 29    & 0.004 & 0.001 \\
Loaded\_Language                  & 10,388 & 1.298 & 0.507 \\
Name\_Calling-Labeling            & 2,012  & 0.252 & 0.098 \\
no\_technique                     & 2,966  & 0.371 & 0.145 \\
Obfuscation-Vagueness-Confusion   & 756   & 0.095 & 0.037 \\
Questioning\_the\_Reputation      & 776   & 0.097 & 0.038 \\
Red\_Herring                      & 50    & 0.006 & 0.002 \\
Repetition                        & 166   & 0.021 & 0.008 \\
Slogans                           & 144   & 0.018 & 0.007 \\
Straw\_Man                        & 25    & 0.003 & 0.001 \\
Whataboutism                      & 28    & 0.004 & 0.001\\
\midrule
\textbf{Total} & \textbf{20,487}  &  & \\ 
\bottomrule
\end{tabular}%
}
\vspace{-0.2cm}
\caption{Distributions of techniques in \arpro. F: Number of spans per technique. P-R: Ratio at the paragraph level. T-R: Ratio at the dataset level.}
\label{tab:data_split_tech_dist}
\vspace{-0.45cm}
\end{table}

\noindent\textbf{Co-occurrence of Techniques.}
To understand the relationship between different techniques, we computed their co-occurrence. In Table \ref{tab: techniques_cooccurrence}, we report the top ten pairs of techniques. This shows that the technique \textit{loaded language} is highly associated with several techniques, including \textit{name calling}, \textit{exaggeration minimization}, and \textit{questioning the reputation}. The strong association between \textit{questioning the reputation} and \textit{name calling labeling} reflects the fact that while the former technique is used to question the reputation of an individual, organization, or entity, the latter emphasizes the statement or message further.  

\begin{table}[]
\centering
\setlength{\tabcolsep}{4pt}
\resizebox{\columnwidth}{!}{%
\begin{tabular}{@{}llr@{}}
\toprule
\multicolumn{1}{c}{\textbf{Technique 1}} & \multicolumn{1}{c}{\textbf{Technique 2}} & \multicolumn{1}{c}{\textbf{Freq.}} \\ \midrule
Loaded\_Language & Name\_Calling-Labeling & 777 \\
Loaded\_Language & Exaggeration-Minimisation & 627 \\
Questioning\_the\_Reputation & Loaded\_Language & 397 \\
Obfuscation-Vagueness-Confusion & Loaded\_Language & 356 \\
Loaded\_Language & Causal\_Oversimplification & 215 \\
Name\_Calling-Labeling & Exaggeration-Minimisation & 205 \\
Loaded\_Language & Doubt & 173 \\
Questioning\_the\_Reputation & Name\_Calling-Labeling & 170 \\
Loaded\_Language & Flag\_Waving & 134 \\
Loaded\_Language & Appeal\_to\_Authority & 134 \\ \bottomrule
\end{tabular}%
}
\caption{Top ten most frequent techniques and their co-occurrence frequency.}
\label{tab: techniques_cooccurrence}
\vspace{-0.45cm}
\end{table}





\section{Experimental Setup}
\label{sec:experiments}
In this study, we aim to establish strong baselines on our \arpro{} dataset to encourage and support the development of models for propaganda detection in text. Our experiments also target the evaluation of the strongest closed LLM to-date, GPT-4, for the task at hand. In this section, we describe the setup and design of experiments to achieve these goals.   

\subsection{Task Formulation}
\label{ssec:tasks}
The task of propaganda detection in text has been formulated covering different classification granularities. It ranged from modeling the problem as a binary classification task (propaganda vs. non-propaganda)~\cite{AAAI2019:proppy}, to a sequence tagging task where the aim is to extract spans of text containing persuasion techniques aiming to influence readers~\cite{EMNLP19DaSanMartino,propaganda-detection:WANLP2022-overview,piskorski2023news,piskorski2023semeval}. Propaganda and persuasion are closely-related, as both aim at influencing readers and both employ persuasion strategies~\cite{jowett2018propaganda}. As previously done in literature, we use these terms interchangeably in this work. We model the task into four granularities, to cover those commonly-observed in literature as follows:
\begin{itemize}[itemsep=0pt, parsep=0pt]
    \item Binary propaganda detection (Binary): Given a text snippet, detect whether it uses any propaganda technique or not. 
    \item Coarse-grained propaganda detection (Multilabel): Given a text snippet, detect the high-level categories of propaganda techniques it contains from the six categories in our 2-tier annotation taxonomy.
    \item Propaganda techniques detection (Multilabel): Given a text snippet, identify propaganda techniques it contains from the 23 persuasion techniques. 
    \item Propaganda text spans identification (Span/Sequence Tagging): Given a text snippet, identify the propaganda techniques it contains and text spans in which these techniques are being used. 
\end{itemize}

Given our dataset includes annotations at the finest granularity (span-level), we created three more versions of the dataset by  mapping these annotations into each of the granularities explained above. 

\subsection{Data Splits}
\label{sec:data_splits}
We split the dataset in a stratified manner\cite{sechidis2011stratification}, allocating 75\%, 8.5\%, and 16.5\% for training, development, and testing, respectively. During the stratified sampling, the multilabel setting was considered when splitting the dataset. This ensures that persuasion techniques are similarly distributed across the splits. 

In Tables \ref{tab:binary_coarse_label_dist} and \ref{tab:data_split_span}, we report the distribution of different data splits for binary, coarse, and span-level labels. Reporting the sequence distribution for multilabel annotations would lead to a large table, which we have omitted in this paper. 

\begin{table}[]
\centering
\setlength{\tabcolsep}{4pt}
\resizebox{0.75\columnwidth}{!}{%
\begin{tabular}{@{}lrrr@{}}
\toprule
\textbf{Label} & \textbf{Train} & \textbf{Dev} & \textbf{Test} \\ \midrule
\textbf{Binary} \\ \midrule
Propagandistic & 3,777 & 425 & 832 \\
Non-Propagandistic & 2,225 & 247 & 494 \\ \cmidrule{2-4}
\textbf{Total} & \textbf{6,002} & \textbf{672} & \textbf{1,326} \\  \midrule
\multicolumn{4}{c}{\textbf{Coarse-grained labels}} \\  \midrule
Call & 176 & 21 & 40 \\
Distraction & 74 & 9 & 16 \\
Justification & 471 & 48 & 102 \\
Manipulative\_Wording & 3,460 & 387 & 757 \\
no\_technique & 2,225 & 247 & 494 \\
Reputation & 1,404 & 163 & 314 \\
Simplification & 384 & 42 & 82 \\ \cmidrule{2-4}
\textbf{Total} & \textbf{8,194} & \textbf{917} & \textbf{1,805} \\ \bottomrule
\end{tabular}%
}
\vspace{-0.2cm}
\caption{Binary and coarse grained label distribution.}
\label{tab:binary_coarse_label_dist}
\vspace{-0.45cm}
\end{table}



\begin{table}[]
\centering
\resizebox{0.9\columnwidth}{!}{%
\begin{tabular}{@{}lrrr@{}}
\toprule
Technique & \multicolumn{1}{l}{Train} & \multicolumn{1}{l}{Dev} & \multicolumn{1}{l}{Test} \\ \midrule
Appeal\_to\_Authority & 192 & 22 & 42 \\
Appeal\_to\_Fear-Prejudice & 93 & 11 & 21 \\
Appeal\_to\_Hypocrisy & 82 & 9 & 17 \\
Appeal\_to\_Popularity & 44 & 4 & 8 \\
Appeal\_to\_Time & 52 & 6 & 12 \\
Appeal\_to\_Values & 38 & 5 & 9 \\
Causal\_Oversimplification & 289 & 33 & 67 \\
Consequential\_Oversimplification & 81 & 10 & 19 \\
Conversation\_Killer & 53 & 6 & 13 \\
Doubt & 227 & 27 & 49 \\
Exaggeration-Minimisation & 967 & 113 & 210 \\
False\_Dilemma-No\_Choice & 60 & 6 & 13 \\
Flag\_Waving & 174 & 22 & 41 \\
Guilt\_by\_Association & 22 & 2 & 5 \\
Loaded\_Language & 7,862 & 856 & 1670 \\
Name\_Calling-Labeling & 1,526 & 158 & 328 \\
no\_technique & 2,225 & 247 & 494 \\
Obfuscation-Vagueness-Confusion & 562 & 62 & 132 \\
Questioning\_the\_Reputation & 587 & 58 & 131 \\
Red\_Herring & 38 & 4 & 8 \\
Repetition & 123 & 13 & 30 \\
Slogans & 101 & 19 & 24 \\
Straw\_Man & 19 & 2 & 4 \\
Whataboutism & 20 & 4 & 4 \\ \midrule
\textbf{Total} & \textbf{15,437} & \textbf{1,699} & \textbf{3,351} \\ \bottomrule
\end{tabular}%
}
\vspace{-0.2cm}
\caption{Distribution of the techniques in different data splits at the span level.}
\label{tab:data_split_span}
\vspace{-0.4cm}
\end{table}

\subsection{Models}
\label{sec:models}
\noindent\textbf{Pe-trained transformer models (PLMs).} These models have showed significant performance gains in diverse NLP tasks. We explored different models to evaluate their performance for our tasks, including binary, coarse and multilabel classification settings. We used \textbf{AraBERT} \citep{baly2020arabert} and XLM-RoBERTa (\textbf{XLM-r})~\citep{conneau2019unsupervised} and fine-tuned them using task-specific classification head over the training subset.   
We used the transformer toolkit~\cite{Wolf2019HuggingFacesTS} to fine-tune the models. Following the approach of \citet{devlin2018bert}, we fine-tuned each model using default settings over five epochs. We conducted five reruns for each experiment with different random seeds, and report the \textit{average performance} over the reruns on the test subset.

\noindent\textbf{GPT-4.} In addition, our experiments consist of zero-shot learning using GPT-4 \cite{openai2023gpt4}. To ensure reproducibility of the zero-shot experiments, we set the temperature value to zero. We used version 0314 of the GPT-4 model, which was released in June 2023. We chose this model due to its accessibility and superior performance compared to other models, such as GPT-3.5, as reported in~\cite{ahuja2023mega}. 
We specifically designed a prompt for each of the four tasks, which we will release as part of experimental resources. For the experiments, we used the LLMeBench framework \cite{dalvi2023llmebench}.

\noindent\textbf{Random Baseline}
For different tasks, different approaches were followed to compute random baselines. For binary classification, we randomly assigned a label from the two potential labels: [``propagandistic", ``non-propagandistic"] to each test instance. For multilabel classification, where multiple labels from a predefined set are required, both the count and choice of labels were random, and these were then designated to the test instance.

\subsection{Evaluation Measures}
We computed both macro-averaged and micro-averaged F$_1$ scores to evaluate the models' performance. These measures have been frequently used in previous studies when reporting on the performance of propaganda detection tasks \cite{dimitrov2021detecting}. Since the span-level task is a multilabel sequence tagging task, it is evaluated using a modified 
F$_1$ measure that accounts for partial matching between the spans across the gold labels and the predictions~\cite{propaganda-detection:WANLP2022-overview}.  
\section{Results and Discussion}
\label{sec:results}

In Table \ref{tab:results}, we present the performance of different classification settings and models. Based on the Micro-F$_1$ measure, the fine-tuned AraBERT models outperforms in two out of the three tasks, while XLM-r performs well in the multilabel task. All models surpass the random baseline results for all tasks, except for GPT-4's zero-shot performance in the binary task. The zero-shot performance for GPT-4 is relatively lower compared to other models across all tasks settings. As discussed in~\cite{bang2023multitask}, its performance heavily depends on the prompt design, which we plan to study in future research. 

\begin{table}[]
\centering
\setlength{\tabcolsep}{4pt}
\resizebox{0.8\columnwidth}{!}{%
\begin{tabular}{@{}llrr@{}}
\toprule
\textbf{Task} & \textbf{Model} & \textbf{Micro-F$_1$} & \textbf{Macro-F$_1$} \\ \midrule
\multirow{4}{*}{\textbf{Binary}}     & Random & 0.510  & 0.503  \\
& AraBERT& \textbf{0.767}& \textbf{0.750}\\
& XLM-r& 0.627  & 0.386  \\
& GPT-4, 0-shot& 0.488  & 0.457  \\ \hline
\multirow{4}{*}{\textbf{Coarse}}     & Random & 0.215  & 0.161  \\
& AraBERT& \textbf{0.656}& \textbf{0.321}\\
& XLM-r& 0.595  & 0.244  \\
& GPT-4, 0-shot& 0.518  & 0.301  \\ \hline
\multirow{4}{*}{\textbf{Multilabel}} & Random & 0.078  & 0.055\\
& AraBERT& 0.543  & 0.086  \\
& XLM-r& \textbf{0.608}& 0.128  \\
& GPT-4, 0-shot& 0.367  & \textbf{0.168}\\\bottomrule
\end{tabular}%
}
\vspace{-0.2cm}
\caption{Results on \arpro{} test set in different classification settings and models. Best per task per measure is \textbf{boldfaced}.}
\label{tab:results}
\vspace{-0.1cm}
\end{table}

\noindent{\textbf{\textit{How effective is GPT-4 for detecting and labeling propagandistic spans in text?}}} To answer this question, we run GPT-4 in zero-shot setting over the testing split of \arpro. We also investigate its performance over six other languages as follows. The model is applied to development subsets\footnote{Gold labels for the testing subsets are not publicly available.} of a recently-released multilingual dataset for the task as part of SemEval23 shared task 3~\cite{piskorski2023semeval}. The dataset covers six languages: English, French, German, Italian, Polish, and Russian. It includes sentences annotated by persuasion spans and the same 23 techniques targeted in this work.

In Table~\ref{tab:gpt_results}, we report the results for the span detection task, which reveals interesting observations. First, results clearly show that GPT-4's performance is really low for this sequence tagging and multilabel task. This is especially noticeable with lower-resourced languages like Polish and Russian. Moreover, putting these results in context of GPT-4's performance on the less fine-grained tasks, referring back to Table~\ref{tab:results}, results are the lowest for the span extraction task. Nevertheless, the Micro-F$_1$ scores observed are significantly higher than those for a random baseline we created, that randomly assigns propaganda techniques to random spans of text in a paragraph~\cite{propaganda-detection:WANLP2022-overview}. 

\noindent{\textbf{Effect of Prompt Design.}} A challenge we faced when prompting GPT-4 to extract spans from text, was the design of the prompt. With such complex task, the model is required to not only return the propaganda techniques, but also the text spans matching these techniques. Since a span might occur multiple times in a paragraph, with different context and propagandistic load, the model should also specify start and end indices of these spans. In our experiments, we observed that although GPT-4 can correctly provide labels and extract associated span texts, it was generating indices not matching the identified spans. This might be interpreted as if the model approached the task as subtasks and generated outputs for them independently. For this work, we apply a post-prediction heuristic to overcome this problem, by assigning for each predicted span, the start and end indices of its first occurrence in a paragraph. 

\begin{table}[]
\centering
\setlength{\tabcolsep}{4pt}
\resizebox{0.9\columnwidth}{!}{%
\begin{tabular}{llrr}
\toprule
\textbf{Lang.} & \textbf{\# Samples} & \multicolumn{1}{c}{\textbf{Micro-F$_1$}} & \multicolumn{1}{c}{\textbf{Macro-F$_1$}} \\ \midrule
Arabic  & 1,326 & 0.118 (0.010)     & 0.130 (0.010)     \\
English & 3,127 & 0.112 (0.008)     & 0.116 (0.006)     \\
French  & 610   & 0.112 (0.015)     & 0.119 (0.013)     \\
German  & 522   & 0.047 (0.011)     & 0.096 (0.006)     \\
Italian & 882   & 0.098 (0.014)     & 0.101 (0.009)     \\
Polish  & 800   & 0.052 (0.007)     & 0.080 (0.004)     \\
Russian & 515   & 0.057 (0.009)     & 0.078 (0.004)    \\\bottomrule
\end{tabular}%
}
\vspace{-0.2cm}
\caption{GPT-4 performance in propaganda span extraction over \arpro{} test set (Arabic), and development sets from SemEval23 shared task 3. Numbers in parentheses indicate performance of the random baseline.}
\label{tab:gpt_results}
\vspace{-0.4cm}
\end{table}
\section{Related Work}
\label{sec:related_work}

\subsection{Computational Propaganda}

Computational propaganda is defined as the use of automated methods and online platforms to intentionally spread misleading information~\cite{woolley2018computational}. It frequently employs various types of content (e.g., fake news and doctored images) across different media platforms, often using tools like bots. The information is typically distributed in various modalities, such as textual, visual, or multi-modal. To limit the effect of propagandistic content in online media, there have been research efforts to develop resources and tools in order to identify and debunk them. Below, we discuss relevant resources and studies that primarily focus on propaganda detection tasks.


\subsection{Existing Resources} 
The majority of research in propaganda detection has primarily centered on the analysis of textual content~\cite{BARRONCEDENO20191849,rashkin-EtAl:2017:EMNLP2017,EMNLP19DaSanMartino,da2020survey,piskorski2023semeval}. Recently several initiatives attempted towards addressing multimodal content such as memes~\cite{SemEval2021-6-Dimitrov}. The development of the \rashkincorpus~is an earlier effort, which utilized distant supervision, meaning articles from a specific news outlet were uniformly labeled based on that outlet's characterization~\cite{rashkin-EtAl:2017:EMNLP2017}. The annotations include \emph{trusted}, \emph{satire}, \emph{hoax}, and \emph{propaganda}. The dataset incorporated articles from the English Gigaword corpus along with content from seven other less reliable news sources, two of which were identified as propagandistic. \citet{BARRONCEDENO20191849} developed~\proppycorpus~corpus, which is labeled as either propaganda or non-propaganda. They conducted experiments on both the \rashkincorpus~and \proppycorpus~datasets. For the \rashkincorpus,~they binarized the labels, distinguishing between propaganda and the other three categories. \citet{Habernal.et.al.2017.EMNLP,Habernal2018b} developed a corpus comprising $1.3K$ arguments annotated for five fallacies. These include \textit{ad hominem}, \textit{red herring}, and \textit{irrelevant authority}, all of which are directly related to propaganda techniques. 

Recent efforts began to stress the importance of fine-grained analysis of specific propagandistic techniques. \citet{EMNLP19DaSanMartino} identified 18 distinct propaganda techniques and created a dataset by annotating news articles based on these techniques. Annotations were done at the fragment level, focusing on two main tasks: (\emph{i}) binary classification — determining whether any of the 18 techniques were employed in a given sentence of an article; and (\emph{ii}) multi-label multi-class classification and span detection — pinpointing specific text fragments that utilized a propaganda technique and identifying the specific technique used. Building on this work, they designed a multi-granular deep neural network that extracts span from the sentence-level task, thereby enhancing the accuracy of the fragment-level classifier. Focusing on the annotation schema, \citet{blodgett2023combined} proposed 23 top-level techniques, of which 10 match the techniques proposed by~\citet{EMNLP19DaSanMartino}.
\citet{piskorski2023semeval} proposed an extension of those techniques and introduced a dataset in multiple languages. Based on this dataset, some studies such as the work of ~\citet{hasanain-etal-2023-qcri} demonstrated that multilingual pre-trained models significantly surpass monolingual models, even in languages not previously seen.

Research on propaganda detection in tweets is somewhat limited due to the scarcity of annotated datasets. Addressing this gap,~\citet{vijayaraghavan-vosoughi-2022-tweetspin} introduced a corpus of tweets with weak labels for fine-grained propaganda techniques. This study also proposed an end-to-end Transformer-based model enhanced with a multi-view approach that integrates context, relational data, and external knowledge into the representations. Focusing on Arabic social media, ~\citet{propaganda-detection:WANLP2022-overview} developed an annotated dataset consisting of 950 tweets. Very recently, another dataset has been released as a part of the ArAIEval shared task covering tweets and news paragraphs annotated in a multilabel setting\footnote{\url{https://araieval.gitlab.io/task1/}}.

Table~\ref{tab:existing_data_sources} summarizes existing datasets specifically developed for the detection of propaganda and techniques. Compared to prior datasets, ours is the largest in terms of number of paragraphs for a particular language.  

\begin{table}[]
\centering
\resizebox{\columnwidth}{!}{%
\begin{tabular}{@{}lllrl@{}}
\toprule
\multicolumn{1}{c}{\textbf{Reference}} & \multicolumn{1}{c}{\textbf{Lang}} & \multicolumn{1}{c}{\textbf{Content}} & \multicolumn{1}{c}{\textbf{\# Items}} & \multicolumn{1}{c}{\textbf{\# T}} \\ \midrule
\rowcolor{gray!25}
~\cite{BARRONCEDENO20191849} & En & News article & 51,000 & 2 \\
\rowcolor{white}
~\cite{EMNLP19DaSanMartino} & En & News article & 451 & 18 \\ 
\rowcolor{blue!25}
~\cite{SemEval2021-6-Dimitrov} & En & Memes & 950 & 22 \\
\rowcolor{lime!25}
~\cite{vijayaraghavan-vosoughi-2022-tweetspin} & En & Tweets & 1,000  & 19 \\
 \rowcolor{white}
~\cite{piskorski2023semeval} & \makecell{En, Fr,\\de, It,\\Pl, Ru, \\Es, El,\\ Ka} & News article & 2,049 & 23 \\
\rowcolor{white}
\rowcolor{green!25}
~\cite{propaganda-detection:WANLP2022-overview} & Ar & & 930 & 19 \\
ArAIEval & Ar & \makecell{Paragraphs,\\Tweets} & 3,189 & 23 \\ 
\rowcolor{magenta!25} \midrule
\textbf{Ours} & Ar & Paragraphs & 8,000 & 23 \\ \bottomrule
\end{tabular}%
}
\caption{Prior datasets for propaganda detection tasks. \# T: Number of techniques/labels.}
\label{tab:existing_data_sources}
\vspace{-0.4cm}
\end{table}

\section{Conclusion and Future Work}
\label{sec:conclusion}
 
In this study, we introduce a large, manually annotated dataset for detecting propaganda techniques in Arabic text. We have collected and annotated $8K$ news paragraphs sourced from $2.8K$ news articles using 23 propaganda techniques. To facilitate future annotation efforts over Arabic text, we constructed Arabic annotation guidelines and release them to the community. Our work provides an in-depth analysis of the dataset, shedding some light on propaganda use in Arabic news media. We examine the performance of various pre-trained models, including GPT-4, across different classification settings targeting four formulations of the propaganda detection task. Our results indicate that fine-tuned models significantly outperform the GPT-4 in zero-shot setting. The experiments also demonstrated that GPT-4 struggled with the task of detecting propagandistic spans from text in seven languages. 

In future work, we plan to explore the correlation between propaganda and other phenomena in news reporting like misinformation. We also plan to extend our work to designing more sophisticated propaganda spans detection models, in addition to investigating the potential of large language models in various zero-shot and few-shot settings to better understand their capabilities.


%

%

\section*{Ethics and Broader Impact}

We collected news articles from various sources and selected specific paragraphs for the annotation. It is important to note that annotations are subjective, inevitably introducing biases into our dataset. However, our clear annotation schema and instructions aim to minimize these biases. We urge researchers and users of this dataset to remain careful of its potential limitations when developing models or conducting further research. Models developed using this dataset could be invaluable to fact-checkers, journalists, social media platforms, and policymakers.

\section*{Acknowledgments}
The contributions of this work were funded by the NPRP grant 14C-0916-210015, which is provided by the Qatar National Research Fund (a member of Qatar Foundation).

\bibliographystyle{lrec-coling2024-natbib}
\bibliography{bib/all}

\begin{thebibliography}{41}
\expandafter\ifx\csname natexlab\endcsname\relax\def\natexlab#1{#1}\fi

\bibitem[{Abdelali et~al.(2023)Abdelali, Mubarak, Chowdhury, Hasanain, Mousi, Boughorbel, Kheir, Izham, Dalvi, Hawasly, Nazar, Elshahawy, Ali, Durrani, Milic-Frayling, and Alam}]{abdelali2023benchmarking}
Ahmed Abdelali, Hamdy Mubarak, Shammur~Absar Chowdhury, Maram Hasanain, Basel Mousi, Sabri Boughorbel, Yassine~El Kheir, Daniel Izham, Fahim Dalvi, Majd Hawasly, Nizi Nazar, Yousseif Elshahawy, Ahmed Ali, Nadir Durrani, Natasa Milic-Frayling, and Firoj Alam. 2023.
\newblock \href {http://arxiv.org/abs/2305.14982} {Benchmarking arabic ai with large language models}.

\bibitem[{Ahuja et~al.(2023)Ahuja, Hada, Ochieng, Jain, Diddee, Maina, Ganu, Segal, Axmed, Bali et~al.}]{ahuja2023mega}
Kabir Ahuja, Rishav Hada, Millicent Ochieng, Prachi Jain, Harshita Diddee, Samuel Maina, Tanuja Ganu, Sameer Segal, Maxamed Axmed, Kalika Bali, et~al. 2023.
\newblock {MEGA}: Multilingual evaluation of generative ai.
\newblock \emph{arXiv preprint arXiv:2303.12528}.

\bibitem[{Alam et~al.(2022{\natexlab{a}})Alam, Cresci, Chakraborty, Silvestri, Dimitrov, Martino, Shaar, Firooz, and Nakov}]{alam-etal-2022-survey}
Firoj Alam, Stefano Cresci, Tanmoy Chakraborty, Fabrizio Silvestri, Dimiter Dimitrov, Giovanni Da~San Martino, Shaden Shaar, Hamed Firooz, and Preslav Nakov. 2022{\natexlab{a}}.
\newblock A survey on multimodal disinformation detection.
\newblock In \emph{Proceedings of the 29th International Conference on Computational Linguistics}, pages 6625--6643, Gyeongju, Republic of Korea. International Committee on Computational Linguistics.

\bibitem[{Alam et~al.(2022{\natexlab{b}})Alam, Mubarak, Zaghouani, Nakov, and Da~San~Martino}]{propaganda-detection:WANLP2022-overview}
Firoj Alam, Hamdy Mubarak, Wajdi Zaghouani, Preslav Nakov, and Giovanni Da~San~Martino. 2022{\natexlab{b}}.
\newblock Overview of the {WANLP} 2022 shared task on propaganda detection in {A}rabic.
\newblock In \emph{Proceedings of the Seventh Arabic Natural Language Processing Workshop}, WANLP~'22, Abu Dhabi, UAE.

\bibitem[{Antoun et~al.(2020)Antoun, Baly, and Hajj}]{baly2020arabert}
Wissam Antoun, Fady Baly, and Hazem Hajj. 2020.
\newblock {AraBERT}: Transformer-based model for {Arabic} language understanding.
\newblock In \emph{Proceedings of the 4th Workshop on Open-Source Arabic Corpora and Processing Tools, with a Shared Task on Offensive Language Detection}, pages 9--15.

\bibitem[{Artstein and Poesio(2008)}]{artstein2008inter}
Ron Artstein and Massimo Poesio. 2008.
\newblock Inter-coder agreement for computational linguistics.
\newblock \emph{Computational Linguistics}, 34(4):555--596.

\bibitem[{Bang et~al.(2023)Bang, Cahyawijaya, Lee, Dai, Su, Wilie, Lovenia, Ji, Yu, Chung et~al.}]{bang2023multitask}
Yejin Bang, Samuel Cahyawijaya, Nayeon Lee, Wenliang Dai, Dan Su, Bryan Wilie, Holy Lovenia, Ziwei Ji, Tiezheng Yu, Willy Chung, et~al. 2023.
\newblock A multitask, multilingual, multimodal evaluation of {C}hat{GPT} on reasoning, hallucination, and interactivity.
\newblock \emph{arXiv preprint arXiv:2302.04023}.

\bibitem[{Barr\'on-Cede{\~n}o et~al.(2019)Barr\'on-Cede{\~n}o, Da~San~Martino, Jaradat, and Nakov}]{AAAI2019:proppy}
Alberto Barr\'on-Cede{\~n}o, Giovanni Da~San~Martino, Israa Jaradat, and Preslav Nakov. 2019.
\newblock Proppy: A system to unmask propaganda in online news.
\newblock In \emph{Proceedings of the Thirty-Third AAAI Conference on Artificial Intelligence (AAAI'19)}, AAAI'19, Honolulu, HI.

\bibitem[{Barr{\'o}n-Cedeno et~al.(2019)Barr{\'o}n-Cedeno, Jaradat, Da~San~Martino, and Nakov}]{BARRONCEDENO20191849}
Alberto Barr{\'o}n-Cedeno, Israa Jaradat, Giovanni Da~San~Martino, and Preslav Nakov. 2019.
\newblock Proppy: Organizing the news based on their propagandistic content.
\newblock \emph{Information Processing \& Management}, 56(5):1849--1864.

\bibitem[{Blodgett et~al.(2023)Blodgett, Bonial, Hudson, and Voss}]{blodgett2023combined}
Austin Blodgett, Claire Bonial, Taylor Hudson, and Clare Voss. 2023.
\newblock Combined annotations of misinformation, propaganda, and fallacies identified robustly and explainably (campfire).

\bibitem[{Chen et~al.(2023)Chen, Zhao, Piao, Ding, and Cui}]{chen2023multimodal}
Pengyuan Chen, Lei Zhao, Yangheran Piao, Hongwei Ding, and Xiaohui Cui. 2023.
\newblock Multimodal visual-textual object graph attention network for propaganda detection in memes.
\newblock \emph{Multimedia Tools and Applications}, pages 1--16.

\bibitem[{Conneau et~al.(2020)Conneau, Khandelwal, Goyal, Chaudhary, Wenzek, Guzm{\'a}n, Grave, Ott, Zettlemoyer, and Stoyanov}]{conneau2019unsupervised}
Alexis Conneau, Kartikay Khandelwal, Naman Goyal, Vishrav Chaudhary, Guillaume Wenzek, Francisco Guzm{\'a}n, Edouard Grave, Myle Ott, Luke Zettlemoyer, and Veselin Stoyanov. 2020.
\newblock \href {https://doi.org/10.18653/v1/2020.acl-main.747} {Unsupervised cross-lingual representation learning at scale}.
\newblock In \emph{Proceedings of the 58th Annual Meeting of the Association for Computational Linguistics}, ACL~'20, pages 8440--8451, Online. Association for Computational Linguistics.

\bibitem[{Da~San~Martino et~al.(2019)Da~San~Martino, Yu, Barr\'{o}n-Cede\~no, Petrov, and Nakov}]{EMNLP19DaSanMartino}
Giovanni Da~San~Martino, Seunghak Yu, Alberto Barr\'{o}n-Cede\~no, Rostislav Petrov, and Preslav Nakov. 2019.
\newblock Fine-grained analysis of propaganda in news articles.
\newblock In \emph{Proceedings of the 2019 Conference on Empirical Methods in Natural Language Processing and 9th International Joint Conference on Natural Language Processing, EMNLP-IJCNLP 2019, Hong Kong, China, November 3-7, 2019}, EMNLP-IJCNLP 2019, Hong Kong, China.

\bibitem[{Dalvi et~al.(2023)Dalvi, Hasanain, Boughorbel, Mousi, Abdaljalil, Nazar, Abdelali, Chowdhury, Mubarak, Ali, Hawasly, Durrani, and Alam}]{dalvi2023llmebench}
Fahim Dalvi, Maram Hasanain, Sabri Boughorbel, Basel Mousi, Samir Abdaljalil, Nizi Nazar, Ahmed Abdelali, Shammur~Absar Chowdhury, Hamdy Mubarak, Ahmed Ali, Majd Hawasly, Nadir Durrani, and Firoj Alam. 2023.
\newblock \href {http://arxiv.org/abs/2308.04945} {{LLMeBench}: A flexible framework for accelerating llms benchmarking}.

\bibitem[{Devlin et~al.(2019)Devlin, Chang, Lee, and Toutanova}]{devlin2018bert}
Jacob Devlin, Ming-Wei Chang, Kenton Lee, and Kristina Toutanova. 2019.
\newblock \href {https://doi.org/10.18653/v1/N19-1423} {{BERT}: Pre-training of deep bidirectional transformers for language understanding}.
\newblock In \emph{{Proceedings of the 2019 Conference of the North {A}merican Chapter of the Association for Computational Linguistics: Human Language Technologies}}, NAACL-HLT~'19, pages 4171--4186, Minneapolis, Minnesota, USA. Association for Computational Linguistics.

\bibitem[{Dimitrov et~al.(2021{\natexlab{a}})Dimitrov, Ali, Shaar, Alam, Silvestri, Firooz, Nakov, and Martino}]{dimitrov2021detecting}
Dimitar Dimitrov, Bishr~Bin Ali, Shaden Shaar, Firoj Alam, Fabrizio Silvestri, Hamed Firooz, Preslav Nakov, and Giovanni Da~San Martino. 2021{\natexlab{a}}.
\newblock Detecting propaganda techniques in memes.
\newblock \emph{arXiv preprint arXiv:2109.08013}.

\bibitem[{Dimitrov et~al.(2021{\natexlab{b}})Dimitrov, Bin~Ali, Shaar, Alam, Silvestri, Firooz, Nakov, and Da~San~Martino}]{SemEval2021-6-Dimitrov}
Dimitar Dimitrov, Bishr Bin~Ali, Shaden Shaar, Firoj Alam, Fabrizio Silvestri, Hamed Firooz, Preslav Nakov, and Giovanni Da~San~Martino. 2021{\natexlab{b}}.
\newblock {S}em{E}val-2021 task 6: Detection of persuasion techniques in texts and images.
\newblock In \emph{Proceedings of the 15th International Workshop on Semantic Evaluation (SemEval-2021)}, pages 70--98, Online. Association for Computational Linguistics.

\bibitem[{Habernal et~al.(2017)Habernal, Hannemann, Pollak, Klamm, Pauli, and Gurevych}]{Habernal.et.al.2017.EMNLP}
Ivan Habernal, Raffael Hannemann, Christian Pollak, Christopher Klamm, Patrick Pauli, and Iryna Gurevych. 2017.
\newblock Argotario: Computational argumentation meets serious games.
\newblock In \emph{Proceedings of the 2017 Conference on Empirical Methods in Natural Language Processing: System Demonstrations}, pages 7--12, Copenhagen, Denmark. Association for Computational Linguistics.

\bibitem[{Habernal et~al.(2018)Habernal, Pauli, and Gurevych}]{Habernal2018b}
Ivan Habernal, Patrick Pauli, and Iryna Gurevych. 2018.
\newblock {Adapting Serious Game for Fallacious Argumentation to German: Pitfalls, Insights, and Best Practices}.
\newblock In \emph{Proceedings of the Eleventh International Conference on Language Resources and Evaluation (LREC 2018)}, pages 3329--3335.

\bibitem[{Hasanain et~al.(2023)Hasanain, El-Shangiti, Nandi, Nakov, and Alam}]{hasanain-etal-2023-qcri}
Maram Hasanain, Ahmed El-Shangiti, Rabindra~Nath Nandi, Preslav Nakov, and Firoj Alam. 2023.
\newblock {QCRI} at {S}em{E}val-2023 task 3: News genre, framing and persuasion techniques detection using multilingual models.
\newblock In \emph{Proceedings of the 17th International Workshop on Semantic Evaluation (SemEval-2023)}, pages 1237--1244, Toronto, Canada. Association for Computational Linguistics.

\bibitem[{{Institute for Propaganda Analysis}(1938)}]{InstituteforPropagandaAnalysis1938}
{Institute for Propaganda Analysis}. 1938.
\newblock {How to Detect Propaganda}.
\newblock In \emph{Propaganda Analysis. Volume I of the Publications of the Institute for Propaganda Analysis}, chapter~2, pages 210--218. New York, NY.

\bibitem[{Jowett and O'donnell(2018)}]{jowett2018propaganda}
Garth~S Jowett and Victoria O'donnell. 2018.
\newblock \emph{Propaganda \& persuasion}.
\newblock Sage publications.

\bibitem[{Landis and Koch(1977)}]{landis1977measurement}
J~Richard Landis and Gary~G Koch. 1977.
\newblock The measurement of observer agreement for categorical data.
\newblock \emph{biometrics}, pages 159--174.

\bibitem[{Liang et~al.(2022)Liang, Bommasani, Lee, Tsipras, Soylu, Yasunaga, Zhang, Narayanan, Wu, Kumar et~al.}]{liang2022holistic}
Percy Liang, Rishi Bommasani, Tony Lee, Dimitris Tsipras, Dilara Soylu, Michihiro Yasunaga, Yian Zhang, Deepak Narayanan, Yuhuai Wu, Ananya Kumar, et~al. 2022.
\newblock Holistic evaluation of language models.
\newblock \emph{arXiv preprint arXiv:2211.09110}.

\bibitem[{Martino et~al.(2020)Martino, Cresci, Barrón-Cedeño, Yu, Pietro, and Nakov}]{da2020survey}
Giovanni Da~San Martino, Stefano Cresci, Alberto Barrón-Cedeño, Seunghak Yu, Roberto~Di Pietro, and Preslav Nakov. 2020.
\newblock A survey on computational propaganda detection.
\newblock In \emph{Proceedings of the Twenty-Ninth International Joint Conference on Artificial Intelligence}, IJCAI~'20, pages 4826--4832.

\bibitem[{Mathet(2017)}]{mathet-2017-agreement}
Yann Mathet. 2017.
\newblock \href {https://doi.org/10.1162/COLI_a_00296} {The agreement measure γcat a complement to γ focused on categorization of a continuum}.
\newblock \emph{Computational Linguistics}, 43(3):661--681.

\bibitem[{Mathet et~al.(2015)Mathet, Widl{\"{o}}cher, and M{\'{e}}tivier}]{Mathet2015}
Yann Mathet, Antoine Widl{\"{o}}cher, and Jean-Philippe M{\'{e}}tivier. 2015.
\newblock {The Unified and Holistic Method Gamma ($\gamma$) for Inter-Annotator Agreement Measure and Alignment}.
\newblock \emph{Computational Linguistics}, 41(3):437--479.

\bibitem[{OpenAI(2023)}]{openai2023gpt4}
OpenAI. 2023.
\newblock \href {http://arxiv.org/abs/2303.08774} {{GPT}-4 technical report}.
\newblock Technical report, OpenAI.

\bibitem[{Passonneau(2006)}]{passonneau2006measuring}
Rebecca Passonneau. 2006.
\newblock Measuring agreement on set-valued items ({MASI}) for semantic and pragmatic annotation.
\newblock In \emph{Proceedings of the Fifth International Conference on Language Resources and Evaluation}, LREC~'06, pages 831--836, Genoa, Italy.

\bibitem[{Perrin(2015)}]{perrin2015social}
Andrew Perrin. 2015.
\newblock Social media usage.
\newblock \emph{Pew research center}, pages 52--68.

\bibitem[{Piskorski et~al.(2023{\natexlab{a}})Piskorski, Stefanovitch, Bausier, Faggiani, Linge, Kharazi, Nikolaidis, Teodori, De~Longueville, Doherty, Gonin, Ignat, Kotseva, Mantica, Marcaletti, Rossi, Spadaro, Verile, Da~San~Martino, Alam, and Nakov}]{piskorski2023news}
Jakub Piskorski, Nicolas Stefanovitch, Valerie-Anne Bausier, Nicolo Faggiani, Jens Linge, Sopho Kharazi, Nikolaos Nikolaidis, Giulia Teodori, Bertrand De~Longueville, Brian Doherty, Jason Gonin, Camelia Ignat, Bonka Kotseva, Eleonora Mantica, Lorena Marcaletti, Enrico Rossi, Alessio Spadaro, Marco Verile, Giovanni Da~San~Martino, Firoj Alam, and Preslav Nakov. 2023{\natexlab{a}}.
\newblock News categorization, framing and persuasion techniques: Annotation guidelines.
\newblock Technical report, European Commission Joint Research Centre, Ispra (Italy).

\bibitem[{Piskorski et~al.(2023{\natexlab{b}})Piskorski, Stefanovitch, Da~San~Martino, and Nakov}]{piskorski2023semeval}
Jakub Piskorski, Nicolas Stefanovitch, Giovanni Da~San~Martino, and Preslav Nakov. 2023{\natexlab{b}}.
\newblock \href {https://doi.org/10.18653/v1/2023.semeval-1.317} {{S}em{E}val-2023 task 3: Detecting the category, the framing, and the persuasion techniques in online news in a multi-lingual setup}.
\newblock In \emph{Proceedings of the 17th International Workshop on Semantic Evaluation (SemEval-2023)}, pages 2343--2361, Toronto, Canada. Association for Computational Linguistics.

\bibitem[{Rashkin et~al.(2017)Rashkin, Choi, Jang, Volkova, and Choi}]{rashkin-EtAl:2017:EMNLP2017}
Hannah Rashkin, Eunsol Choi, Jin~Yea Jang, Svitlana Volkova, and Yejin Choi. 2017.
\newblock Truth of varying shades: {A}nalyzing language in fake news and political fact-checking.
\newblock In \emph{Proceedings of the 2017 Conference on Empirical Methods in Natural Language Processing}, pages 2931--2937. Association for Computational Linguistics.

\bibitem[{Roitero et~al.(2020)Roitero, Soprano, Fan, Spina, Mizzaro, and Demartini}]{Roitero2020}
Kevin Roitero, Michael Soprano, Shaoyang Fan, Damiano Spina, Stefano Mizzaro, and Gianluca Demartini. 2020.
\newblock \href {https://doi.org/10.1145/3397271.3401112} {Can the crowd identify misinformation objectively? {T}he effects of judgment scale and assessor's background}.
\newblock In \emph{Proceedings of the 43rd International ACM SIGIR Conference on Research and Development in Information Retrieval}, SIGIR '20, pages 439--448, Virtual Event, China. Association for Computing Machinery.

\bibitem[{Sechidis et~al.(2011)Sechidis, Tsoumakas, and Vlahavas}]{sechidis2011stratification}
Konstantinos Sechidis, Grigorios Tsoumakas, and Ioannis Vlahavas. 2011.
\newblock \href {https://link.springer.com/chapter/10.1007/978-3-642-23808-6_10} {On the stratification of multi-label data}.
\newblock In \emph{Machine Learning and Knowledge Discovery in Databases}, ECML-PKDD~'11, pages 145--158, Berlin, Heidelberg. Springer Berlin Heidelberg.

\bibitem[{Sharma et~al.(2022)Sharma, Alam, Akhtar, Dimitrov, Da~San~Martino, Firooz, Halevy, Silvestri, Nakov, and Chakraborty}]{ijcai2022p781}
Shivam Sharma, Firoj Alam, Md.~Shad Akhtar, Dimitar Dimitrov, Giovanni Da~San~Martino, Hamed Firooz, Alon Halevy, Fabrizio Silvestri, Preslav Nakov, and Tanmoy Chakraborty. 2022.
\newblock Detecting and understanding harmful memes: A survey.
\newblock In \emph{Proceedings of the Thirty-First International Joint Conference on Artificial Intelligence}, IJCAI~'22, pages 5597--5606, Vienna, Austria. International Joint Conferences on Artificial Intelligence Organization.
\newblock Survey Track.

\bibitem[{Sheikh~Ali et~al.(2021)Sheikh~Ali, Mansour, Elsayed, and Al-Ali}]{ali2021arafacts}
Zien Sheikh~Ali, Watheq Mansour, Tamer Elsayed, and Abdulaziz Al-Ali. 2021.
\newblock {AraFacts}: the first large arabic dataset of naturally occurring claims.
\newblock In \emph{Proceedings of the Sixth Arabic Natural Language Processing Workshop}, pages 231--236.

\bibitem[{Vamanu(2019)}]{vamanu-2019}
Iulian Vamanu. 2019.
\newblock \href {https://doi.org/doi:10.1515/opis-2019-0014} {Fake news and propaganda: A critical discourse research perspective}.
\newblock \emph{Open Information Science}, 3(1):197--208.

\bibitem[{Vijayaraghavan and Vosoughi(2022)}]{vijayaraghavan-vosoughi-2022-tweetspin}
Prashanth Vijayaraghavan and Soroush Vosoughi. 2022.
\newblock \href {https://doi.org/10.18653/v1/2022.naacl-main.251} {{TWEETSPIN}: Fine-grained propaganda detection in social media using multi-view representations}.
\newblock In \emph{Proceedings of the 2022 Conference of the North American Chapter of the Association for Computational Linguistics: Human Language Technologies}, pages 3433--3448, Seattle, United States. Association for Computational Linguistics.

\bibitem[{Wolf et~al.(2020)Wolf, Debut, Sanh, Chaumond, Delangue, Moi, Cistac, Rault, Louf, Funtowicz, Davison, Shleifer, von Platen, Ma, Jernite, Plu, Xu, Le~Scao, Gugger, Drame, Lhoest, and Rush}]{Wolf2019HuggingFacesTS}
Thomas Wolf, Lysandre Debut, Victor Sanh, Julien Chaumond, Clement Delangue, Anthony Moi, Pierric Cistac, Tim Rault, Remi Louf, Morgan Funtowicz, Joe Davison, Sam Shleifer, Patrick von Platen, Clara Ma, Yacine Jernite, Julien Plu, Canwen Xu, Teven Le~Scao, Sylvain Gugger, Mariama Drame, Quentin Lhoest, and Alexander Rush. 2020.
\newblock \href {https://doi.org/10.18653/v1/2020.emnlp-demos.6} {Transformers: State-of-the-art natural language processing}.
\newblock In \emph{Proceedings of the 2020 Conference on Empirical Methods in Natural Language Processing: System Demonstrations}, EMNLP~'20, pages 38--45, Online. Association for Computational Linguistics.

\bibitem[{Woolley and Howard(2018)}]{woolley2018computational}
Samuel~C Woolley and Philip~N Howard. 2018.
\newblock \emph{Computational propaganda: political parties, politicians, and political manipulation on social media}.
\newblock Oxford University Press.

\end{thebibliography}



\appendix
\section{Appendix}
\label{sec:append-how-prod}

\section{Annotation Platform}
\label{sec:app_annotation_platform}

\begin{figure*}
    \centering
    \includegraphics[width=1\linewidth]{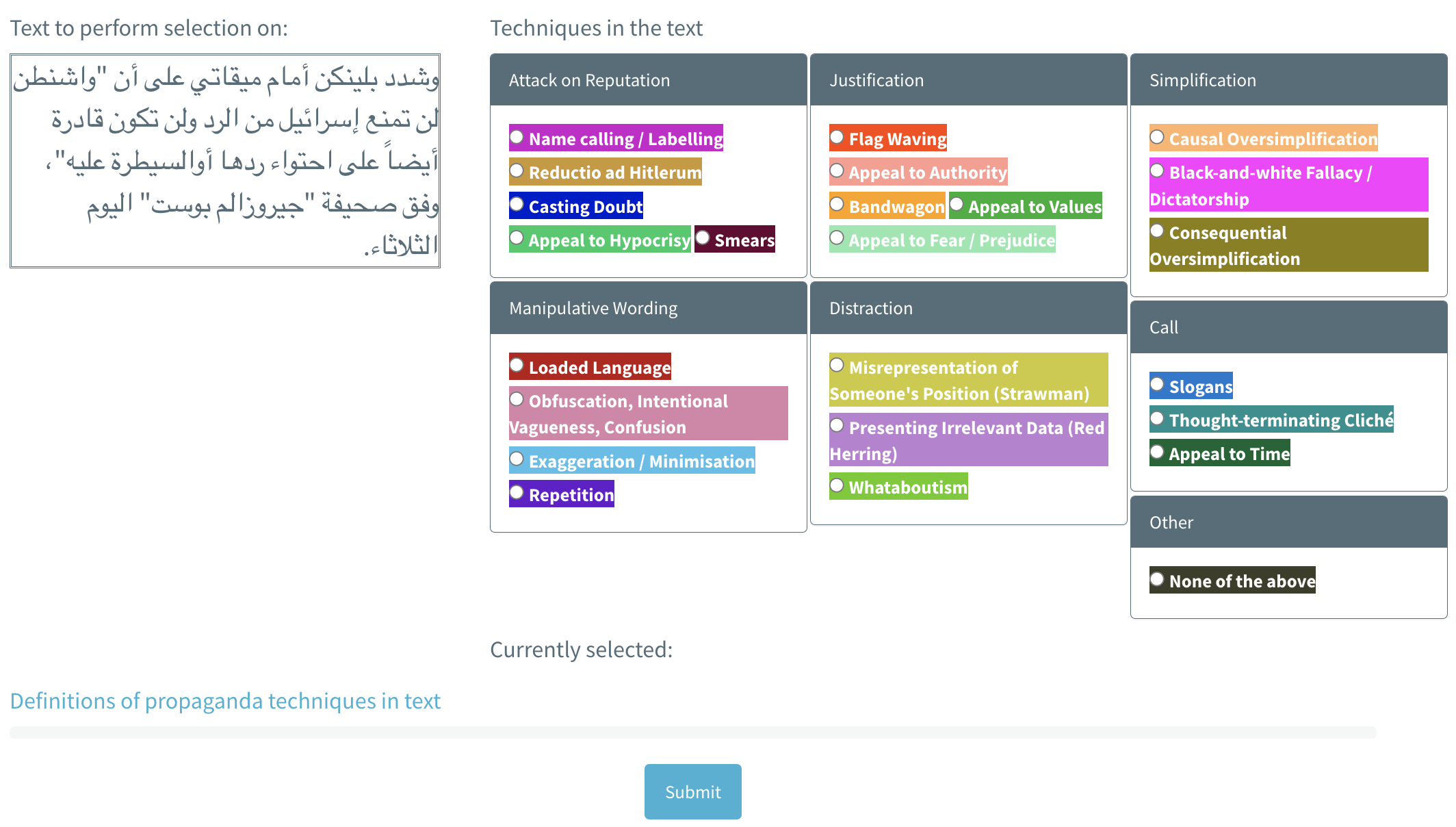}
    \caption{A screenshot of the annotation platform for \textbf{phase 1 (annotation)}.}
    \label{fig:example_of_annotation_interface}
\end{figure*}

\begin{figure*}
    \centering
    \includegraphics[width=1\linewidth]{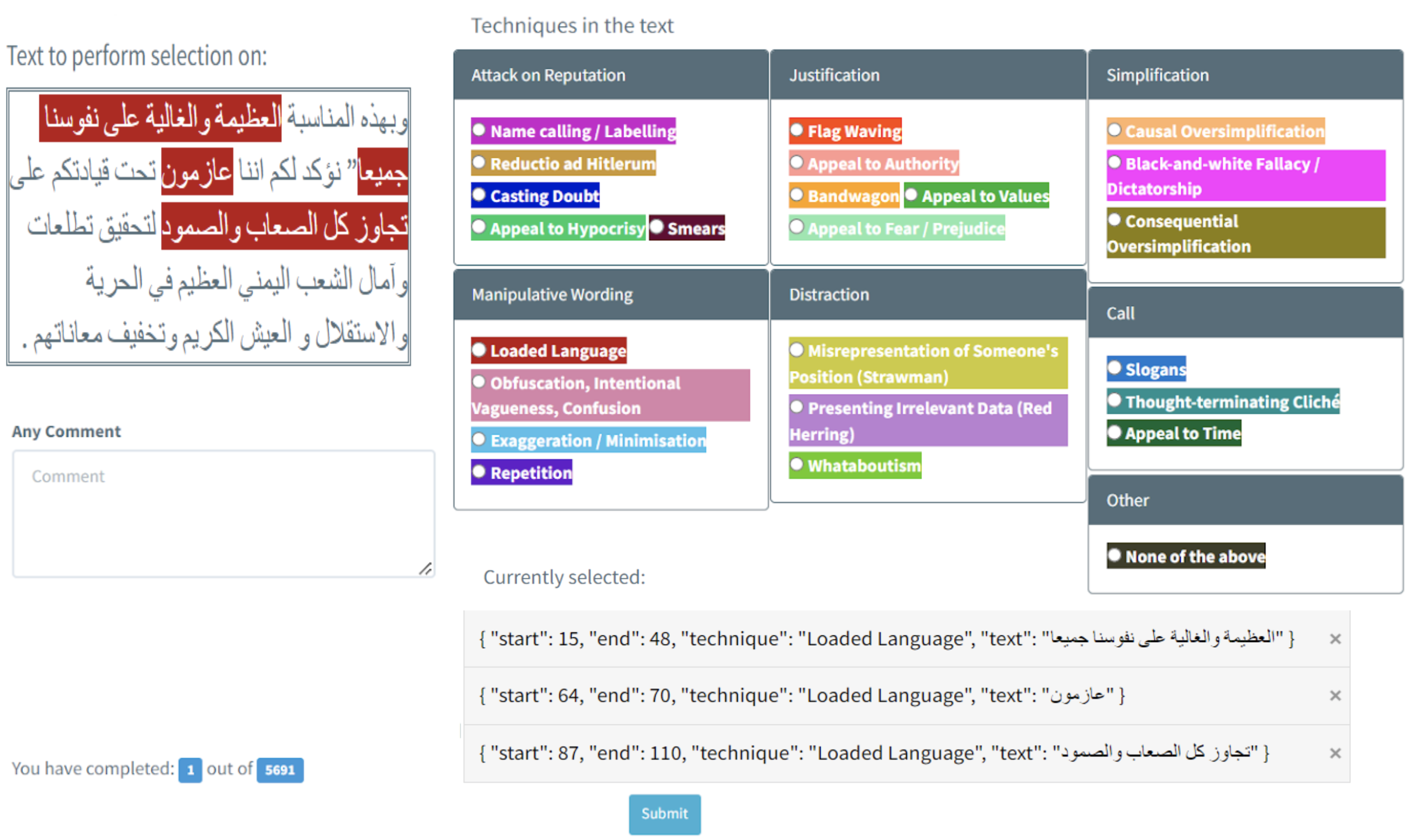}
    \caption{A screenshot of the annotation platform for \textbf{phase 2 (consolidation)}.}
    \label{fig:example_of_consolidation_interface}
\end{figure*}

In Figure \ref{fig:example_of_annotation_interface} and \ref{fig:example_of_consolidation_interface}, we present the annotation interfaces for phase 1 and phase 2, respectively. The annotations in the phase 2 interface are pre-filled with those completed in phase 1, and consolidators have the option to remove existing annotations or add new ones. 

\section{Annotators Training}
\label{sec:app_annotation_training}

The training process consisted of the following steps:
\begin{enumerate} 
\item We provided annotation guidelines to both teams. In the annotation guidelines, each propagandistic technique was supplemented with examples from previously annotated data. Annotators had the ability to refer to these guidelines whenever needed during the annotation process.
\item We setup pilot annotations as training exercises for the annotators. Once the annotations were completed, the data was collected and evaluated for quality.
\item Based on the results from step 2, meetings were organized to address and discuss the challenges annotators faced. These challenges could be related to interpreting the guidelines or issues encountered with the annotation platform.
\end{enumerate}
\newpage


\section*{Supplementary material}

4. Steps 2 and 3 were repeated iteratively to ensure continuous improvement in annotation quality.

## D. Annotation Guidelines

### (1) التسميات أو التصنيفات
### (1) Name calling/Labeling

شكل من أشكال المحاجَجة أو الجدال توجه أو تطلق خلاله التسميات أو الصفات المشحونة على فرد أو مجموعة، وعادة ما توجه بشكل مهين أو تحقيري. تُستَقى هذه الأوصاف أو التسميات من أشياء يخشاها الجمهور المستهدف، أو يكرهها، أو يجدها غير مرغوب بها أو غير محبوبة، أو على العكس قد تستقى من صفات مرغوبة ومحبوبة، وتحاول هذه الطريقة أن تدفع القارئ ليحكم على ذات أو جوهر الكيان موضع الجدل دون النظر إلى أي حقائق عنه. وهو أسلوب صياغة متلاعب بالقارئ، حيث يعمل على التأثير عليه بتغيير جزء من الجملة بتضمينها صفة أو مجموعة من الصفات بدلاً من استخدام حجة كاملة متكونة من فرضية ونتيجة. فمثلاً ما في الخطاب السياسي، عادةً ما تستخدم مثل هذه الصفات والأسماء كإشارات ترمز للتوجه السياسي والآراء والصفات الشخصية والارتباط ببعض المنظمات وكذلك الإهانات.

الفرق الرئيسي بين هذا الأسلوب وأسلوب **"اللغة المشحونة"** (المعرف في النقطة 20) هو أن أسلوب **"التسميات"** يركز فقط على وصف أو تصنيف الكيان المستهدف.

**ما الذي يجب أن تحدده في النص:** الاسم أو الصفة التي استخدمت في النص بغرض الإساءة مع علامات التنصيص إن وجدت.

أمثلة

- الإرهابي"بشار" ينعت الثوار ضده بـ"الثيران" ويشيد بقطعان "الموالين".
- الناطق باسم كتائب القسام "للصهاينة": احشدوا ما شئتم فقد أعددنا لكم أصنافاً من الموت ستجعلكم تلعنون أنفسكم.

### (2) الذنب بالتداعي (الحجة النازية)
### (2) Reductio ad hitlerum

مهاجمة خصم أو نشاط ما عن طريق ربطه بشخص أو مجموعة أو مفهوم آخر له دلالات سلبية قوية لدى الجمهور المستهدف، وأكثر مثال شائع على هذا الأسلوب هو مقارنة كيان ما مع النظام النازي وهتلر، وهذا سبب تسميته بأسلوب "الحجة النازية"، لكن هذا الأسلوب لا يقتصر على المقارنة بمجموعة النازية. قد يستخدم هذا الأسلوب بادعاء وجود رابط أو علاقة مماثلة بين الكيان المستهدف وأي فرد أو مجموعة أو حدث في الحاضر أو الماضي، بحيث يكون لدى الجمهور المستهدف تصور سلبي مؤكد عنه، مثلاً: اعتباره شخص أو شيء فاشل، أو تم تصويره بهذه الطريقة السلبية.

**ما الذي يجب أن تحدده في النص:** الجزء الذي يشير إلى كيان ما (شخص أو مؤسسة مثلاً) بالإضافة إلى ذكر الكيان الآخر (الذي يعتبر شريراً أو سلبياً)، وقد قام بالنشاط أو العمل ذاته أو نشاط مماثل له، ويبدو من النص أن هذا النشاط سلبي. قد يُذكَر النشاط الذي يقوم به الكيان المستهدف بشكل ضمني دون التصريح به.

أمثلة

- وزير الخارجية التركي مولود تشاووش أوغلو قال أننا لانصف الحكومة الألمانية الحالية بالنازية، ولكن أفعالهم تذكرنا بتلك الحقبة.

- يعبر عن افتتان ابن سلمان بطريقة الزعيم النازي للتخلص من خصومه وأعدائه.
- أنت تحب اللون الأسود مثل الإرهابيين. إذن أنت إرهابي!
- يتكلم هذا الشخص كأحد أعضاء مجلس النواب.

**ملاحظة:** الفرق الرئيسي بين أسلوب **"التسميات"** (المعرف في النقطة 1) و أسلوب **"الذنب بالتداعي"** هو أن الأول يقصد الإهانة أو المدح باستخدام مصطلح أو تسمية معينة فقط، دون الإشارة إلى أي فكرة أو عمل آخر، على سبيل المثال: الرئيس "النازي" يستخدم أسلوب **"التسميات"**. في حين أن الأسلوب الثاني يربط بوضوح بين فكرة أو فعل للشخص المستهدف بشخص آخر يفعل شيئًا مشابهًا، فنقول مثلاً: لديه خطة تذكرنا بخطة النازيين.

### (3) إثارة الشكوك
### (3) Casting Doubt

إثارة الشكوك حول الصفات الشخصية لشخص أو شيء ما بهدف التشكيك بمصداقيته بشكل عام أو جودته بدلاً من استخدام حجة مناسبة تتعلق بالموضوع، مثلاً قد يذكر الخلفية المهنية للكيان المستهدف كوسيلة لتشويه حجته أو التشكيك بمصداقيتها. يمكن أيضاً استثارة الشكوك عبر الإشارة إلى بعض الأفعال أو المناسبات التي نفذها الكيان المستهدف أو خطط لها ولم تنجح، أو يبدو -على الأرجح- أنها لن تؤدي إلى تحقيق الأهداف المخطط لها.

**ما الذي يجب أن تحدده في النص:** يجب أن تكتفي بتحديد الجزء من النص الذي يشكك في المصداقية والكيان (الشخص أو المؤسسة) الذي يتم التشكيك في مصداقيته. لا داعي لشمل السياق الكامل للموضوع أثناء التحديد.

أمثلة

- تهاوي "قبة إسرائيل الحديدية" أمام صواريخ المقاومة يدعو للتساؤل ما مدى فعاليتها أمام هجوم متطور ؟
- وأكدت مديرة العمليات بمنظمة العفو الدولية في فرنسا ناتالي غودار، في إشارة إلى استقبال الأوكرانيين أن "ما رأيناه في الأسابيع الأخيرة يتناقض بشدة مع الطريقة التي تحدثت بها السلطات الرسمية العام الماضي عن استقبال الأفغان الفارين من بلادهم ."

### (4) الاتهام بالنفاق ("أنت تفعل ذلك أيضاً")
### (4) Appeal to Hypocrisy

مهاجمة سمعة الكيان المستهدف باتهامه بالنفاق أو التناقض بشكل مباشر أو ضمنيًا بتوضيح التناقضات بين المواقف أو الأفعال المختلفة التي اتخذها أو قام بها هذا الكيان في الماضي. وإحدى الأساليب الخاصة لإظهار نفاق الكيان تتمثل باتهام الشخص المستهدف بالقيام بنفس الفعل الذي ينتقدك عليه الآن، ومن هنا جاءت تسمية هذه التقنية: "أنت تفعل ذلك أيضاً".

**ما الذي يجب أن تحدده في النص:** العبارات أو الجمل التي تقارن بين فعلين: فعل حالي وفعل سابق قام به الشخص أو الكيان المستهدف، واستُحضِر ذلك الفعل السابق لاتهام ذلك الكيان بالنفاق.

أمثلة

- انتقدت منظمة العفو الدولية سياسة المعايير المزدوجة التي تعتمدها فرنسا في التعامل مع اللاجئين من خلال إغلاق الأبواب بوجه الأفغان وفتحها للأوكرانيين .

العرق والجنس والحزب السياسي وغيرها من الانتماءات والجماعات، والحجج التي قد يستخدمها هذا الأسلوب تكون غير مبنية على المنطق، بل تكون مبنية على افتراضات ناتجة من الآراء الشخصية أو التحيزات السابقة تجاه الموضوع، ويمكن القول أن هذا الأسلوب يحاول استثارة مشاعر الجمهور بدلاً من مخاطبة المنطق للتلاعب بالجمهور وكسب الجدال، أو قد يحاول ترديد مشاعر الجمهور لفتح المجال للمزيد من الجدال.

**ما الذي يجب أن تحدده في النص**: الجزء الذي يشير إلى القيم الوطنية/الرموز، بالإضافة إلى الفعل أو الفكرة التي يتم الترويج والدعوة لها في حال ذكرت في النص.

**أمثلة**

- قال سامي الجميّل: معركتنا لبنانية ولا يمكننا إنقاذ طائفة بمعزل عن طائفة أخرى لذا علينا خلق جبهة عابرة للطوائف .
- الدخول في هذه الحرب سيمنحنا مستقبلاً أفضل في بلدنا .

### (7) الاحتكام إلى سُلطة (الإذعان للسلطة)
### (7) Appeal to Authority

يعطي هذا الأسلوب وزناً لحجة أو فكرة أو معلومة ما عبر التصريح بأنها صادرة عن كيان (شخص أو منظمة) يُعتَبَر ذا سلطة. قد يكون الكيان المذكور ذا سلطة أو خبرة وقدرات فعلية تؤهله للحديث في المجال المحدد، ولكن ما يميز هذا الأسلوب عن مجرد استقاء أو اقتباس ونقل المعلومات عن مصدر ما هو أن النبرة المستخدمة أثناء الكتابة تدل على محاولة النص التنفع من وزن أو أهمية الكيان صاحب السلطة المزعومة لتبرير معلومة أو ادعاء أو استنتاج معين. يجب أن يشمل في هذا التصنيف النصوص التي تحاول إقناع القارئ بشيء ما عبر الاحتكام إلى خبرة أو سلطة كيان ما سواء كانت هذه الخبرات حقيقية أو لا. وتشمل إحدى الحالات الخاصة لهذا الأسلوب الإشارة إلى كاتب النص ذاته (أو المؤسسة التي يمثلها) كمرجعية ذات سلطة.

**ما الذي يجب أن تحدده في النص**: الجزء الذي يشير إلى السلطة أو الخبير وبعض من كلامه أو رأيه أو عمله إن وجد، والاستنتاج أو الفكرة المبنية على رأيه أو كلامه في حال ذكرت في النص.

**أمثلة**

- يقول ريتشارد دوكينز، عالم الأحياء التطوري و الخبير الأول في هذا المجال، أن عملية التطور حقيقية ولذا هي حقيقة يجب تقبلها.
- الشمس تدور حول الأرض لأن القرآن يقول ذلك بوضوح لا لبس فيه.
- يؤكد العالِم الكبير وليم جينكينز الحائز على نوبل في الفيزياء أنه سوف يقضى على فيروس الإنفلونزا بجميع أنواعه بحلول عام ألفين وخمسين، ومثل هذا العالِم الفذ لا يُستهان برأيه.

### (8) الاحتكام إلى عامة الناس (الاحتكام إلى الكثرة)
### (8) Bandwagon

يعطي هذا الأسلوب وزناً لحجة أو فكرة معينة عبر الادعاء أن "الجميع" (أو الغالبية العظمى) يوافقون عليها، أو أن "لا أحد" يختلف معها. وبهذا يُشَجَّع الجمهور على تبني نفس الفكرة بشكل جماعي، وعلى القيام بالفعل المذكور في النص بالنظر إلى "الجميع" كسلطة ذات تأثير. قد يشير مصطلح "الجميع" إلى عامة الناس أو الكيانات الرئيسية والجهات الفاعلة في مجال معين أو البلدان، إلخ. وتبعاً لذات المنطق السابق، قد يحاول هذا الأسلوب إقناع الجمهور بعدم فعل شيء ما لأن "لا أحد آخر يقوم بنفس الفعل".

---

- بدت قطر في كثير من الأحيان وكأنها دولة مختلفة عن باقي المنطقة،
- إذ تستضيف حركة حماس، رغم أنها أقامت في السابق أيضا بعض العلاقات التجارية مع إسرائيل
- برر القائمون على الحملة ضد مونديال قطر 2022 أهداف حملتهم بقلقهم من الآثار المترتبة على تبريد ملاعب كأس العالم، فقالوا إنها ستلحق ضررا كبيرا بالأرض والبيئة، وفي الوقت ذاته تجاهل هؤلاء ذكر أي آثار سلبية على البيئة نتيجة قيام فرنسا بتدفئة ملاعبها بالكهرباء.

**ملاحظة**: هناك علاقة بين هذا الأسلوب وأسلوب **"تغيير الموضوع"** (المذكور في النقطة 13)، ولكن أسلوب **"تغيير الموضوع"** يركز على تحويل الانتباه عن الموضوع بدلاً من مهاجمة الخصم بشكل مباشر. ويمكننا أن نعتبر أسلوب **"الاتهام بالنفاق"** كحالة خاصة من أسلوب **"إثارة الشكوك"**.

### (5) التشكيك بالسمعة (تشويه السمعة  تسميم البئر)
### (5) Smears

يُستخدَم هذا الأسلوب لمهاجمة سمعة الكيان المستهدف عبر إثارة ادعاءات سلبية جداً عنه، مع التركيز بشكل خاص على تقويض شخصيته أو مكانته الأخلاقية بدلاً من الاعتماد على حجة متعلقة بموضوع النقاش الحالي. وقد تكون هذه الادعاءات صحيحة أو خاطئة ولكن هذا لا يؤثر على فعالية هذا الأسلوب في الإقناع، وقد تستخدم الألفاظ الملوثة للسمعة في أي مرحلة أثناء النقاش. ومن الممكن أن تمارس هذا الأسلوب عن طريق التشكيك بسمعة أو مصداقية الخصم بشكل استباقي قبل أن تتوفر له الفرصة للتعبير عن نفسه، مما يسبب انحيازاً في تصورات الجمهور تجاهه، ومن هنا جاءت إحدى تسميات هذا الأسلوب: "تسميم البئر".

**ما الذي يجب أن تحدده في النص**: العبارات أو الجمل التي تذكر شيئاً سلبياً حول الشخص أو الكيان المستهدف.

**أمثلة**

- لدى خصمي سجل من الكذب ومحاولات سابقة لتغطية تعاملاته غير النزيهة بابتسامة لطيفة . لا تدعه يقنعك بتصديق كلماته.
- قبل أن نبدأ في النقاش، أود أن أقول أني كثيراً ما تمنيت أن يفهم الرجال حق المرأة في الإجهاض، غير أنهم بحكم موقعهم الذكوري لا يستطيعون رؤية الموضوع من منظور المرأة.

**ملاحظة**: الفرق الرئيسي بين هذا الأسلوب و أسلوب **"إثارة الشكوك"** (المعرف في النقطة 3) هو أن أسلوب إثارة الشكوك يركز على التشكيك في القدرات والمصداقية بشكل خاص، بينما يهدف أسلوب **"التشكيك بالسمعة"** إلى تقويض السمعة العامة والصفات الأخلاقية والسلوك وما إلى ذلك. يمكننا اعتبار هذا الأسلوب كحالة خاصة من أسلوب **"إثارة الشكوك"**.

### (6) التذرُّع بالوطنية (التلويح بالعلم)
### (6) Flag-waving

تبرير فكرة أو الترويج لها باستثارة افتخار الجمهور بانتمائهم إلى مجموعة أو كيان ما أو تسليط الضوء على الفوائد التي تستفيد منها تلك المجموعة. وتعد استثارة الانتماء الوطني مثالاً واضحاً ومعتاداً لهذا الأسلوب ومن هنا جاء اسمه. لكن الانتماء الوطني ليس الانتماء الوحيد الذي قد يستدعيه هذا الأسلوب، فقد يركز على أي انتماء آخر مثل

- قد يمر الاقتصاد الألماني بأكبر انكماش في تاريخه إن أُوقف استيراد الغاز الروسي من دون إيجاد بدائل.
- تشكو كارول أبي كرم -نقيبة مصانع الأدوية والأمصال- من أن 12 مصنعا بلبنان بحاجة لنحو مليون ليتر من المازوت شهريا، "مما يعني أن رفع الدعم عن المحروقات سيضاعف الفاتورة الدوائية".
- وعندما سئل عما إذا كان يتعين على الأوروبيين الخوف من اللاجئين القادمين من سوريا، قال الأسد إنهم خليط، مشيراً إلى أن أغلبيتهم من "السوريين الشرفاء الوطنيين". لكنه أضاف أن هناك إرهابيين تسللوا بينهم وأن هذا أمر حقيقي.

يجب أن نوضح في هذه المرحلة أن الأساليب من 6 إلى 10 تسعى لإقناع الجمهور المستهدف بفعل أو فكرة ما بتقديم مبررات موافقة أو مخالفة لهذا الفعل. لتصنيف النص على أنه يحتوي على أحد أساليب التبرير (أساليب 6 إلى 10)، يجب أن يحتوي النص على جزئين: 1) بيان وذكر صريح أو ضمني لفكرة أو فعل يجب دعمها أو القيام بها، أو يجب فعل أو دعم ما يخالفها، 2) مبرر للفعل أو الفكرة (أو ضدها) ويعتمد هذا المبرر على اللعب على وتر القيم أو القومية أو الانتشار والشعبية أو الخوف، إلخ. وعلى سبيل المثال، ليس كل نص يتكلم عن الخوف أو يذكر شيئاً مخيفاً يعد تابعاً للأسلوب رقم 10 "**إثارة الخوف**". لتوضيح هذا بشكل أكبر، نذكر هنا المثالين التاليين:

1. قال بوتين:" على روسيا أن تنفذ عملية عسكرية خاصة في أوكرانيا، وإلا فإن القوات الأوكرانية النازية ستغزو روسيا وقد خططت بالفعل لتدمير وقصف بعض المدن ..."
2. يستخدم بوتين الخوف لتبرير العملية العسكرية الخاصة في أوكرانيا.

يجب أن يُصنَّف النص الأول على أنه يتبع أسلوب **"إثارة الخوف"** (بوتين يستدعي الخوف لدى جمهوره لتبرير أفعاله (الحرب تجاه أوكرانيا)، بينما يجب تصنيف المثال الثاني على أنه يتبع أسلوب **"التشكيك بالسمعة"** (ينتقد كاتب النص بوتين بسبب لجوئه لإثارة الخوف لدى الجمهور في سبيل دعم أفعاله). من الضروري فهم هذه الفروق الدقيقة بين الأساليب المختلفة لتقدر على تصنيف النصوص بشكل صحيح.

### (11) مهاجمة رجل القش
### (11) Misrepresentation of Someone's Position (Straw Man)

يعطي هذا الأسلوب انطباعاً بأن النص فند حجة أو نظرية الخصم، ولكنه في الواقع لم يتعرض لحجة الخصم الحقيقية إنما قام باستبدالها بأخرى مختلفة. في هذا الأسلوب، تُنشَئ أولاً حجة جديدة من خلال استبدال الحجة الأصلية بشيء مرتبطاً بها يبدو إلى حد ما، ولكنه في الواقع نسخة مختلفة أو مشوهة أو مبالغ فيها من الحجة الأصلية. يقوم النص بعدها بتفنيد الحجة البديلة للأصلية. عادة ما تكون الحجة البديلة أسهل بالتفنيد، مما يوهم القارئ بأن النص نجح بدحض الحجة الحقيقية. غالبًا ما تمثل الحجة البديلة إعادة صياغة مسيئة أو تفسيراً مشوهاً لما يقصده الخصم.

**ما الذي يجب أن تحدده في النص:** قد يمتد السياق ذو الصلة عبر أكثر من جملة عند استخدام هذا الأسلوب، لكن يجب أن تحد التظليل بالجزء من النص (جملة أو جزء منها) الذي يتسبب بتشتيت الانتباه عن الحجة الأساسية للخصم.

**أمثلة**

---

**ما الذي يجب أن تحدده في النص:** الجزء الذي يشير إلى شيء (فعل أو قول) يبدو أنه مدعوم بكثرة و/أو له شهرة، بالإضافة إلى الجزء من النص الذي يذكر الفعل أو الفكرة التي يجب تبنيها، أو الاستنتاج الذي وصل إليه النص بناءً على ذلك الفعل أو القول المشهور في حال ذكر في النص، إذ قد تظهر تلك الفكرة أو ذلك الفعل بشكل ضمني.

**أمثلة**

- استطلاعات الرأي تُشير إلى فوز ساحق للحزب الوطني، ولذا ينبغي عليك أن تصوّت للحزب الوطني.
- الصيام المتقطع نظام حمية شائع للغاية وينصح به كثيرون، مشيدين بقدرته على الإسهام في إنقاص الوزن وأنه لا يضر أبداً ويلائم الجميع.
- سيسعى الجميع لشراء الإصدار الجديد لذلك الهاتف الذكي حين توفره في الأيام القادمة.

### (9) الاحتكام إلى القيم
### (9) Appeal to Values

يعطي هذا الأسلوب وزناً لفكرة ما عبر ربطها بالقيم التي يراها الجمهور المستهدف قيمًا إيجابية. ويقدم النص هذه القيم كمرجع موثوق بهدف تدعيم أو نفي حجة أو فكرة ما. ومن الأمثلة على هذه القيم: التقاليد، الدين، الأخلاق، العمر، الإنصاف، الحرية، الديمقراطية، السلام، الشفافية، وما إلى ذلك. يجب التنويه أنه في حال ذكرت هذه القيم خارج سياق حجة منطقية متعلقة بالموضوع واستخدمت كلمات تدل على تلك القيم لوصف شيء أو شخص ما، فإنه يجب تصنيف هذه الكلمات تحت أسلوب آخر وهو أسلوب **"اللغة المشحونة"** (المعرف في النقطة 20).

**ما الذي يجب أن تحدده في النص:** الجزء الذي يشير إلى القيم بالإضافة إلى الجزء من النص الذي يذكر الفعل أو الفكرة التي يجب تبنيها أو الاستنتاج الذي وصل إليه النص بناءً على تلك القيم إذا ذكر ذلك الاستنتاج بشكل مباشر في النص.

**أمثلة**

- العادات والتقاليد السارية تؤكد أن الأصل في المجتمع هو الزواج.
- اعتدنا في شركتنا على منح الموظفين من الرجال رواتباً أعلى من النساء، لذا سنواصل الالتزام بهذا المعيار.

### (10) إثارة الخوف واستدعاء التحيز
### (10) Appeal to Fear/Prejudice

يعمل هذا الأسلوب على الترويج لفكرة ما أو التنفير منها بالاعتماد على نفور الجمهور من هذه الفكرة أو فكرة بديلة لها، مثلاً باستغلال بعض الأحكام المسبقة لديهم تجاه الفكرة أو بديلها، قد تكون الفكرة البديلة هي الوضع الراهن، وفي هذه الحالة يتم وصف الوضع الحالي بطريقة مخيفة باستخدام لغة مشحونة محملة بالمشاعر. وإذا كان الخوف الذي يسعى النص لإثارته مرتبطاً بعواقب قرار ما، فغالباً ما يُجمَع هذا الأسلوب مع أسلوب **"المنحدر الزلق"** (المعرَّف في النقطة 16). وإن ذكر النص بديلين فقط لمواجهة الموقف عندها يتم استخدام هذا الأسلوب مع أسلوب **"المعضلة الزائفة"** (المعرَّف في النقطة 15).

**ما الذي يجب أن تحدده في النص:** الجزء الذي يشير إلى المخاوف أو الانحيازات تجاه فكرة أو فعل ما، بالإضافة إلى الجزء من النص الذي يذكر الفعل أو الفكرة التي يجب تبنيها أو الاستنتاج الذي وصل إليه النص بناءً على عملية التخويف إذا ذكر ذلك الاستنتاج بشكل مباشر في النص.

**أمثلة**

أن تحد التظليل بالجزء من النص (جملة أو جزء منها) الذي يتسبب بتشتيت الانتباه عن الموضوع الأساسي.

**أمثلة**

- قطر تتفق بغزارة على نيمار وليس على محاربة الإرهاب .

- يبدو أن البلديات الفرنسية المشاركة بالحملة ضد ظروف العمال في مونديال قطر نسيت أن متحف الإنسان الذي تحتضنه العاصمة باريس يعرض جماجم مقاومين جزائريين قطعت رؤوسهم أثناء احتلال فرنسا للجزائر .

**تنبيه:** هناك فرق دقيق بين هذا الأسلوب وأسلوب **"التشكيك بالنفاق ("أنت تفعل ذلك أيضاً")"** المعرف في النقطة رقم 4. يركز أسلوب **"أنت تفعل ذلك أيضاً"** على إظهار نفاق الشخص أو معاييره المزدوجة تجاه ذات الموضوع المطروح، أما في أسلوب **"ماذا عن؟"** فتُقَدَّم معلومات غير متعلقة بالموضوع الحالي، وقد يهدف بهذا التدليل بشكل عام على نفاق الشخص أو عدم كفاءته. المثالين التاليين يبينان الفرق بين الأسلوبين:

- "بعد هذا الفشل، يطالبنا الرئيس بالاستقالة، لكنه هو ذاته لن يستقيل" هو مثال على **"أنت تفعل ذلك أيضاً"**

- ما المثال التالي: "هذا الرئيس الذي لم ينتخب حتى، غير قادر على التعامل مع هذه الأزمة " فيتبع أسلوب **"ماذا عن؟"**.

## (14) التبسيط السببي المفرط (السبب الأحادي)
## (14) Causal Oversimplification

يزعم النص المكتوب وجود مسبب واحد لنتيجة معينة، ولكن في الواقع، يوجد عدد من الشروط التي تجتمع لتسبب تلك النتيجة. ويمكن أن نلاحظ هنا مثالين منطقيين أو نمطين يتبعان هذا الأسلوب:

وقع الحدث **ص** بعد **س**، فهذا يعني أن **س** هو المسبب الوحيد لحدوث **ص**.
إذا سبب س وقوع الحدث **ص**، فهذا يعني أن **س** هو العامل الوحيد المسبب لحدوث **ص**.

يفترض المرء أن الأسباب وراء وقوع حدث ما هي: إما أ أو ب أو ج، إلخ، ولا يأخذ بعين الاعتبار أن كلها مجتمعة أو مجموعة منها تسببت بالحدث.

**ما الذي يجب أن تحدده في النص:** أقصر جزء من النص يتبع نمط الاستنتاج المذكور في الأمثلة المنطقية السابقة (المبتدئ بكلمة "فهذا يعني")، أو يذكر استنتاجاً خاطئاً مبنياً على هذا التبسيط. في الكثير من الحالات، لا تظهر كل أجزاء هذا النمط أو الأسلوب في النص بشكل مباشر.

**أمثلة**

- انتشر العنف في المدارس وانخفض الأداء الأكاديمي للطلاب منذ ظهور ألعاب الفيديو التي تحتوي على عنف.
ولهذا، يجب حظر ألعاب الفيديو التي تحتوي على عنف مما سيحسن الوضع في المدارس.

- تعيش مصر أوضاعا اقتصادية صعبة بسبب الاعتماد الكامل على الديون .

- أفادت نساء أن أولياء أمورهن الرجال منعوهن من الدراسة في الخارج أو جامعات مختلطة، مما منعهن من الدراسة واتخاذ حياة مهنية مستقبلية.

- أكدت لوبن خلال مناظرة رئاسية متلفزة تمسكها بفكرتها المثيرة للجدل بشأن فرض قانون لحظر حجاب الرأس الإسلامي في الأماكن العامة مدعية أنه قانون للدفاع عن الحرية. فرد ماكرون "فرنسا، موطن التنوير والكونيّة،
ستصبح أول دولة في العالم تحظر الرموز الدينية في الأماكن العامة ".

- السيد النائب يطلب خفض الزيادات المخصصة للخدمات الصحية بنسبة ٢٠٪، ولكن
كيف لنا أن نبخس الصحة نصيبها من اهتمامنا ونحرم أطفالنا من حقهم في التطعيم والرعاية ؟

## (12) تقديم معلومات خارج الموضوع (الرنجة الحمراء)
## (12) Presenting Irrelevant Data (Red Herring)

محاولة لتحويل الانتباه عن المسألة الرئيسية في الجدل أو النقاش القائم، وذلك بطرح تفاصيل غير هامة أو موضوع أو معلومات مختلفة لا صلة لها بالموضوع المعني. ويهدف الشخص الذي يقوم بتشتيت الموضوع إلى دفع النقاش باتجاه قضية أخرى يمكن أن يرد عليها بشكل أفضل، أو ترك الموضوع الأصلي دون التعامل معه أو الرد عليه. ويُعد أسلوب مهاجمة رجل القش حالة خاصة من هذا الأسلوب نظراً لأنه يسعى لتشتيت الانتباه عن الموضوع الأساسي بطرح حجة بديلة لا تمثل حجة الخصم بشكل دقيق.

**ما الذي يجب أن تحدده في النص:** قد يمتد السياق ذو الصلة عبر أكثر من جملة عند استخدام هذا الأسلوب، إلا أنه يجب أن تكتفي بتظليل الجزء من النص (جملة أو جزء منها) الذي يتسبب بتشتيت الانتباه عن الموضوع الأساسي.

**أمثلة**

- بخصوص علاقتها مع السعودية، وفرت قطر منصة للمعارضين الإسلاميين الذين اعتبرتهم السعودية وحلفاؤها تهديدا، في الوقت الذي تمد فيه علاقات الصداقة مع إيران عدو الرياض
وتستضيف أكبر قاعدة عسكرية أمريكية في المنطقة .

- تعرضت جودة منتجاتنا في الآونة الاخيرة للكثير من الانتقادات،
لذا قررنا إقامة تنزيلات جديدة ليستطيع العملاء شراء المزيد من المنتجات بسعر أقل !

- اتهم ماكرون منافسته بأن لديها رغبة غير منقوصة في إخراج فرنسا من الاتحاد الأوروبي، ردت لوبان
بتعهد بإعادة الأموال إلى جيوب ملايين الفرنسيين الذين زادوا فقرا خلال 5 سنوات من رئاسة ماكرون .

## (13) تغيير الموضوع ("ماذا عن؟")
## (13) Whataboutism

أسلوب يحاول تشويه موقف الخصم عبر اتهامه بالنفاق دون تفنيد حجته بشكل مباشر. فبدلاً من الإجابة على سؤال حرج أو تفنيد حجة منطقية يطرحها الخصم أثناء النقاش، يسعى مُتَّبع هذا الأسلوب للرد بسؤال حَرِج مضاد يمثل اتهاماً مضاداً للخصم، فقد يذكر مثلاً المعايير المزدوجة للخصم. ويهدف مُتَّبع هذا الأسلوب لصرف الانتباه عن الموضوع الأساسي واستبداله بموضوع آخر.

**ما الذي يجب أن تحدده في النص:** قد يمتد السياق ذو الصلة عبر أكثر من جملة عند استخدام هذا الأسلوب، وعلى الرغم من ذلك، يجب

- يجب أن يبقى اختيار المقررات التي تُدرَّس بالجامعات أمرًا متروكًا للأساتذة؛ لأننا

  إذا سمحنا لرغبات الطلبة بالتأثير في هذا الاختيار فسوف يتصورون أنهم يُديرون التعليم، وهذا سيؤدي إلى انهيار النظام، وسرعان ما نجدنا بإزاء جامعات لا تُعلِّم شيئًا.

**أمثلة (مؤيدة لعمل معين):**

- إن توقفنا عن شراء الغاز من روسيا، ستفلس روسيا مما سينهي الحرب في أوكرانيا وبعدها ستبدأ عملية انضمام أوكرانيا إلى الناتو.

- لو تمكن السودان من السيطرة على مورد الذهب فسيحقق عائدات لا تقل عن 5 مليارات دولار في العام و سيكون هذا المبلغ كافياً لتغطية العجز التجاري السنوي في البلاد وبالتالي سيؤدي لتحسين معاش المواطنين وكبح الغلاء الطاحن.

## (17) الشعارات
## (17) Slogans

عبارات موجزة وملفتة للنظر، وقد تتضمن تصنيفًا وتنميطًا. وتميل الشعارات إلى اللعب على وتر العواطف لدى المتلقي.

**ما الذي يجب أن تحدده في النص:** الشعار فقط، لا داعي لتظليل الاستنتاج الناتج عن ذكر هذا الشعار، وإذا أحيط الشعار بعلامات تنصيص، يجب أن تشمل تلك العلامات بالتظليل.

**أمثلة**

- الأستراليون يرفعون شعار " قوتنا فى وحدتنا" لمواجهة أزمة الحرائق، ويقدمون درسا "ملهما" في التعامل مع المأساة. .

- معاً ضد الحرب في اليمن!

- لا أحد يتصدى لإجرامهم، إلا شباب فلسطين بإمكاناتهم القليلة، وبهاماتهم العالية، ويعلنون أن هذه الأرض لنا ، وهذه السماء لنا ، وهذه القدس لنا.

## (18) مُنهي المحادثة (كليشيهات لإيقاف التفكير)
## (18) Thought-terminating cliché

كلمات أو عبارات لا تشجع على التفكير النقدي والمناقشة الهادفة حول موضوع معين. يعد هذا الأسلوب شكلاً من أشكال اللغة المشحونة، وغالبًا ما يتم تمريره كحكمة شعبية. ويهدف هذا الأسلوب إلى إنهاء الجدل ومنع تشكل أي تنافر لدى القارئ بين ما يعتقده وبين الاستنتاج الذي يسعى النص لإقناعه به.

**ما الذي يجب أن تحدده في النص:** أقصر جزء من النص الذي قد يتسبب بإيقاف النقاش أو الحديث.

**أمثلة**

- الشخصية لا تتغير

- الشباب العربي لا يقرأ

- المرأة العربية مسلبة الحقوق

## (15) المعضلة الزائفة ("إما أَبيض أو أسود"، الديكتاتورية)
## (15) Black-and-white Fallacy/Dictatorship

يسمى هذا الأسلوب أحيانًا بمغالطة "إما/ أو"، وهي مغالطة منطقية تقدم خيارين فقط أو وجهتي نظر ممكنتين لا أكثر، بينما هناك خيارات أو وجهات نظر أخرى. في بعض الحالات القصوى، يحدد النص للجمهور التصرفات أو الإجراءات التي يجب اتخاذها، مع القضاء على أي خيارات أخرى ممكنة وهذا ما نقصده هنا بالدكتاتورية. ويتبع هذا الأسلوب مثالاً منطقيًا أو نمطًا معينًا كالتالي:

**"إما أَبيض أو أسود":** أ و ب هما البديلان الوحيدان فقط لحل مشكلة ما، أو القيام بمهمة معينة. ولا يمكن أن يكون الخيار المناسب هو أ. إذا فالخيار ب هو الحل الوحيد.

**"الدكتاتورية":** الحل الوحيد للمشكلة هو الخيار أ.

**ما الذي يجب أن تحدده في النص:** أقصر جزء من النص يتوافق مع الأمثلة المنطقية المذكورة سابقاً. في الكثير من الحالات، لا تظهر كل أجزاء هذا النمط أو الأسلوب في النص بشكل مباشر.

**أمثلة**

- قال ترامب "يتحتم عليكم انتخابي. سواء أعجبكم أم لا، لا يشكل ذلك فرقا لأن بلادنا ستذهب إلى الجحيم إذا جاء أي من هؤلاء الأشخاص "

- علينا أن نُبيحَ الإجهاض دون قيدٍ أو شرط، وإلا فإننا نرغم الأطفال على أن ينشئوا في كنف آباء لا يريدونهم.

## (16) المعضلة التبسيط المفرط للعواقب (المنحدر الزلق)
## (16) Consequential Oversimplification

في هذا الأسلوب، تُرفض حجة أو فكرة مباشرة، وبدلاً من مناقشة ما إذا كانت منطقية و أو صحيحة، يدعي ويؤكد النص وبدون أي دليل أن القبول والتسليم بالحجة أو الفكرة المقدمة يعني قبول أفكار أو حجج أو مقترحات أخرى تعتبر سلبية. الفكرة الأساسية وراء أسلوب "المنحدر الزلق" هو التأكيد على أن وقوع حدث معين أو القيام بفعل ما سيشعل سلسلة أحداث لها بعض الآثار السلبية الكبيرة.

يمكننا تقديم المثال المنطقي أو النمط التالي لهذا الأسلوب:
إذا كان أ سيحدث، فسيتبعه حدوث ب، ج، د حيث:
أ هو أمر يحاول النص رفضه
ب، ج، د تمثّل عواقب سلبية متوقعة في حال حدث أ

وبصورة موازية لما سبق، قد يُستخدم أسلوب "المنحدر الزلق" للتشجيع على فعل معين في سياق بَعِد القارئ بأن القيام بذلك الفعل سيؤدي إلى نتائج إيجابية في النهاية. في هذه الحالة، يمكننا تقديم المثال المنطقي أو النمط التالي:
إذا كان أ سيحدث، فسيتبعه حدوث ب، ج، د حيث:

- أ هو أمر يحاول النص تأييده

- ب، ج، د تمثّل عواقب إيجابية متوقعة في حال حدث أ

**ما الذي يجب أن تحدده في النص:** الجزء من النص الذي يتوافق مع المثال المنطقي أو النمط المذكور سابقاً.

**أمثلة (رافضة لعمل معين):**

- إذا سمحنا اليوم ببعض الضوابط القانونية على الحديث العام أو الكتابة الصحفية، فسوف نسمح غدًا بمزيدٍ من القيود، وهكذا حتى يأتي اليوم الذي نجد أنفسنا فيه نعيش في ظل دولةٍ بوليسية فاشية.

**ما الذي يجب أن تحدده في النص:** أقصر جزء من النص الذي يتسبب بالتشويش أو الحيرة، وقد يتمثل هذا في كلمة واحدة أو قد يمتد لعدة كلمات أو عبارات يجب قراءتها كاملاً لفهم الإرباك الذي تسببه.

**أمثلة**

- مروان فلفال: يجب أن يترفع العمل الحكومي عن الصراع السياسي وينشغل بأمهات القضايا .

- وهذا يشترط، بطبيعة الحال، صفقة سياسية باتجاه تخفيف حدة الاستقطاب القائم بين الفائزين والخاسرين .

- دعا مشروع قانون وزيرة الخارجية الأميركية إلى تقديم تقرير سنوي عن "
- حالات انتقام الحكومات العربية للمدنيين ممّن ينخرطون في علاقات شخصية مع الإسرائيليين ".

## (22) المبالغة أو التقليل
## (22) Exaggeration or Minimisation

الإفراط في تمثيل أو إظهار أو وصف شيء ما، أي جعل الأشياء أكبر أو أفضل أو أسوأ مما هي عليه في الحقيقة، باستخدام عبارات مثل: "الأفضل على الإطلاق" ، "الجودة مضمونة"، أو إظهار الشيء بشكل أقل أهمية أو أصغر مما هو عليه في الواقع، على سبيل المثال: أن يذكر الشخص أن الإهانة التي وجهها لشخص آخر هي "مجرد مزحة".

**ما الذي يجب أن تحدده في النص:** الجزء الذي يقدم وصفاً مبالغاً أو مقلاً من هدف أو شيء أو شخص ما. يجب أن تشمل الكيان المستهدف بالتظليل أيضاً.

**أمثلة**

- لماذا تصارعت معها؟ لم أكن أصارعها. كنا نلعب فقط .

- انسحب الديمقراطيون بمجرد انتهاء خطاب ترامب في محاولة واضحة للإشارة إلى أنهم
- لا يحتملون الوجود في نفس الغرفة مع الرئيس .

- دخل مئات الآلاف من أفراد طواقم التمريض في إنجلترا وويلز وأيرلندا الشمالية في إضراب في الأمس.

## (23) التكرار والتأكيد
## (23) Repetition

حين يلجأ النص لهذا الأسلوب فإنه يَعمَّد لاستخدام نفس الكلمة أو العبارة أو القصة أو الصورة بشكل متكرر آملاً أن يؤدي هذا التكرار إلى إقناع الجمهور.

**ما الذي يجب أن تحدده في النص:** جميع أجزاء النص التي تكرر نفس الرسالة أو المعلومات التي ذكرت في موضع سابق في النص، بالإضافة إلى تظليل أول ظهور لتلك الرسالة المعلومات. إن كان من الصعب تحديد الأجزاء التي يجب تظليلها بالضبط، عند ظهور التكرار في نفس الجملة على سبيل المثال، فيجب تظليل الجملة كاملة في تلك الحالات. من المهم أن نبين أن تكرار شيء ما لا يمكن اعتباره أسلوب إقناع بحد ذاته، فقد يتكرر ذكر شيء ما أو موضوع أو شخص معين ضمن نقاش يستهدف أو يشمل ذلك الشيء بعينه، ومن الطبيعي عندها أن يتكرر ذكر ذلك الشيء في النص. كمثال على ذلك، من المتوقع أن تتكرر كلمة سوريا في مقالة تتحدث عن الوضع السياسي في سوريا ولا يُعدُ هذا أسلوباً للإقناع.

## (19) الضغط بعامل الزمن ("تصرف الآن")
## (19) Appeal to Time

يتمحور هذا الأسلوب حول فكرة أن الوقت قد حان لعمل معين. غالباً ما يُصاغ النص التابع لهذا الأسلوب على نمط "تصرف الآن!"

**ما الذي يجب أن تحدده في النص:** أقصر جزء من النص الذي يحث على القيام بفعل ما باستخدام أسلوب الضغط بعامل الوقت. يجب أن يُظلل الجزء من النص الذي يذكر الزمن أو الوقت بالإضافة إلى الفعل الذي يجب القيام به.

**أمثلة**

- إذا لم يتلقى غالبية السكان اللقاح في غضون شهر، سيقضي علينا الوباء !

- أكد رئيس وفد من منظمة التعاون الإسلامي، حسين إبراهيم طه، اليوم الاثنين، أن
- هذا هو الوقت المناسب لتعبر المنظمة عن صداقتها تجاه روسيا .

- قال وزير الخارجية الأمريكي أنطوني بلينكن في بيان: "قبل عام تقريبا، أطاح الجيش السوداني بالحكومة التي يقودها المدنيون وقوّض التطلعات الديمقراطية للشعب.
- حان الوقت الآن لإنهاء الحكم العسكري ".

## (20) اللغة المشحونة (اللغة المحملة بالمشاعر)
## (20) Loaded Language

استخدام كلمات وعبارات محددة قادرة على استثارة عواطف (سلبية أو إيجابية) قوية لدى المتلقي. ويهدف هذا الأسلوب للتأثير وإقناع الجمهور بأن الحجة المقدمة في النص صحيحة. يُعرَف هذا الأسلوب أيضاً باسم العبارات الملطفة أو المجادلة باستخدام اللغة الانفعالية.

**ما الذي يجب أن تحدده في النص:** نكتفي بتظليل الكلمات أو العبارات المحملة بالمشاعر أو المفخخة، بدون تظليل السياق الذي ظهرت فيه. كقاعدة عامة، لا يجب أن نزيد من طول النص المظلل إلا إن وجدنا أن كل كلمة جديدة نظللها تضيف المزيد من المشاعر أو الحمل الشعوري.

**أمثلة**

- لا يمكن للمتملق أن ينتقد أو يحاسب ولي نعمته ، لأن في اعتقاده أن غرضه الدنيوي في ملك سيده .

- نقلاً عن حركة حماس: ما يجري في المسجد الأقصى مجزرة حقيقية ستدفع سلطات الاحتلال الإسرائيلية ثمنها.

## (21) الإبهام، غموض متعمد، ارتباك، التباس
## (21) Obfuscation, Intentional vagueness, Confusion

تعمد استخدام كلمات غير واضحة أو مبهمة أثناء تقديم الحجة حتى يُكَوِّن المتلقي تفسيراته الخاصة. من الأمثلة على هذا الأسلوب استخدام عبارات غير واضحة ذات تعريفات أو معانٍ متعددة مما يعيق وصول القارئ إلى ذات الاستنتاج المذكور في النص. أي أن هذا العبارات قد تتسبب بالوصول إلى تأويلات مختلفة لذات النص. كما تندرج تحت هذا الأسلوب العبارات غير الدقيقة أو التي تتعمد ألا تجيب بشكل كامل، أو تجيب بشكل غامض على السؤال المطروح.

كالموجه الوحيد لهم أثناء التصنيف. قد يؤدي هذا إلى الإفراط في التصنيف تبعاً لبعض أساليب الإقناع. لقد لاحظنا هذه الظاهرة على وجه الخصوص عند التصنيف تبعاً لأساليب التبسيط (الأساليب 14، 15، 16). ليس كل ما يشبه "التبسيط" يندرج تحت تعريف التبسيط في تصنيف أسلوب الإقناع المطروح.

- كثيراً ما يستخدم مؤلفو المقالات الإخبارية أسلوباً ساخراً أو متهكماً في الكتابة وهذا لا يعد أسلوب إقناع بحد ذاته، إنما يمكن النظر إليه على أن له هدفاً يتوافق مع أحد أهداف أساليب الإقناع المُعرَّفة. غالباً ما يهدف هذا الأسلوب لمهاجمة سمعة شخص أو شيء ما. في مثل هذه الحالات، يجب أن تُصنف النص تبعاً لما تراه الأنسب من أساليب الإقناع المعرفة هنا أو تحت الفئة: "أخرى" إذا لم يندرج تحت أي أسلوب آخر.

- أثناء تظليلك لأسلوب إقناع في جزء من النص، قد تجد أن الأساليب تتداخل أو تتشابك في هذا الجزء وهذا لا بأس به فهذه ظاهرة متكررة. يجب أن تحدد النص التابع لكل واحد من الأساليب الظاهرة في النص حتى لو تشابكت.

- قد تصادف أثناء عملك مقالات لا تحتوي على أي أسلوب إقناع، فلا تضيع وقتك بالبحث عن شيء غير موجود!

- يجب ألا تلجأ إلى أي مصادر خارجية لتقرر ما إذا كنت يجب أن تصنف و تحدد جزءًا ما من النص. يجب أن يعتمد تصنيفك على النص بذاته.

**أمثلة**

- بدأت الأحداث تترى متسارعة بعد انقلاب 3 يوليو/تموز 2013 فانطلقت مظاهرات منددة بالانقلاب وأهله، وأخرى مرحبة مؤيدة، ولكن ميدان رابعة والنهضة وغيرهما من معاقل الرافضين للانقلاب كانت تستحوذ على اهتمام وسائل الإعلام العالمية، وكانت سلطة الانقلاب ضيقة الصدر تتربص بكل من يعارض الانقلاب أو يناوئ الحكم الجديد.

- يعتقد المخرج أن الوقت قد حان لفوز ديب بجائزة الأوسكار . وأكد المخرج، أن "جوائز الأوسكار دائما ما تكون أمرا غامضا، ولكن جوني كان واحدا من أفضل الممثلين خلال فترة طويلة ولم يفز بالأوسكار، ينبغي أن يفوز بها ".

## (24) أخرى
## (24) Other

قد تواجه في بعض الحالات نصاً يبدو أنه كُتب بأسلوب يهدف للإقناع، ولكنه لا يندرج تحت أي من الفئات أو الأساليب السابقة. نرجو تصنيف وتظليل هذا النص باستخدام هذه الفئة.

**ما الذي يجب أن تحدده في النص:** الجزء الذي يبدو أنه يحتوي على أسلوب للإقناع، ولا توجد قاعدة ثابتة وواضحة هنا لما يجب تظليله.

نرجو منك أثناء عملك على تصنيف وتحديد النصوص أن تتبع ما يلي من التعليمات المهمة:

- يجب أن تتبع منهجاً محافظاً حذراً. إن شككت بأن جزءاً من النص يَتَّبع أحد أساليب البروباغاندا ولكنك لست متأكداً، **يجب ألا تقوم بتصنيف أو تظليل هذا الجزء** .

- إذا احترت في طول النص الذي يجب أن تظلله أو تحدده، يجب عليك تحديد الحد الأدنى من النص الذي يدل على أسلوب الإقناع المستخدم.

- نرجو عدم تظليل علامات الترقيم التي تظهر قبل وبعد الجزء من النص الذي تريد تظليله، أما العلامات التي تظهر في داخل هذا الجزء فيجب أن تشملها بالتظليل.

- يجب عليك تجنب التحيز الشخصي وألا تسمح لمشاعرك وآرائك الخاصة المتعلقة بموضوع معين أن تؤثر على عملك في التصنيف لهذا الموضوع!

- نرجو منك الحذر حتى لا تخلط بين المهمة المطلوبة (وهي اكتشاف أساليب الإقناع في النص)، وبين التحقق من صحة الادعاءات، أو تدقيق المعلومات في النص. قد يحتوي جزء من النص على ادعاء صحيح أو دقيق، ولكن هذا لا يعني أو يدل أن هذا النص لا يحتوي على أحد أساليب الإقناع الذي يجب عليك تحديده.

- من المهم جداً قراءة إرشادات التصنيف المُعرَّفة هنا والالتزام بالتعريفات الواردة هنا أثناء التصنيف، نظرًا لأن الحكام الذين يقومون بالتصنيف في هذه المهمة غالبًا ما يميلون لتفسير أسماء أساليب الإقناع حسب وجهة نظرهم، ويستخدمون هذه التفسيرات


\end{document}